%% file: main.tex
\documentclass[]{fairmeta}
\usepackage{microtype}
\usepackage{lmodern}
\usepackage{amsfonts}
\usepackage{graphicx}
\usepackage{booktabs}
\usepackage{float}
\usepackage{wrapfig}
\usepackage{amsmath}
\usepackage[most]{tcolorbox}
\usepackage{amsthm}
\usepackage{algpseudocode}
\usepackage[linesnumbered,lined,boxed,commentsnumbered,ruled,longend]{algorithm2e}
\usepackage[capitalize,noabbrev]{cleveref}
\theoremstyle{plain}

\theoremstyle{definition}

\theoremstyle{remark}

\usepackage{enumitem}
\usepackage{booktabs}
\usepackage{arydshln}

\makeatletter
\def\adl@drawiv#1#2#3{%
  \hskip.5\tabcolsep
  \xleaders#3{#2.5\@tempdimb #1{1}#2.5\@tempdimb}%
  #2\z@ plus1fil minus1fil\relax
  \hskip.5\tabcolsep}
\newcommand{\cdashlinelr}[1]{%
  \noalign{\vskip\aboverulesep
    \global\let\@dashdrawstore\adl@draw
    \global\let\adl@draw\adl@drawiv}
  \cdashline{#1}
  \noalign{\global\let\adl@draw\@dashdrawstore
    \vskip\belowrulesep}}
\makeatother

\title{\raisebox{-0.25ex}{\includegraphics[width=0.08\linewidth]{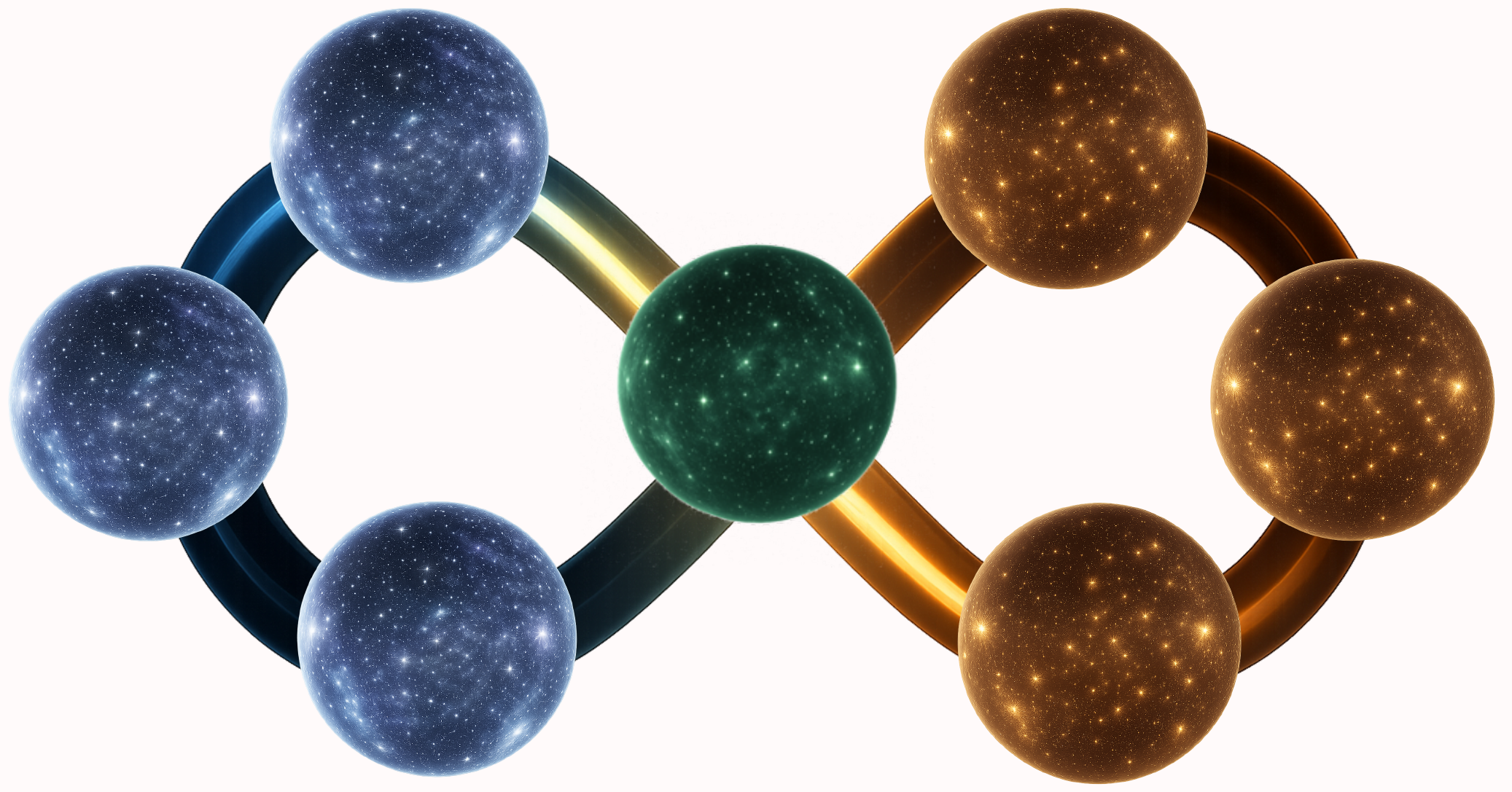}} \fontsize{20pt}{22pt}\selectfont Multiverse: Your Language Models Secretly Decide How to Parallelize and Merge Generation}

\author[*\dagger]{Xinyu Yang}
\author[*\dagger]{Yuwei An}
\author[\dagger]{Hongyi Liu}
\author[\dagger \ddagger] {Tianqi Chen}
\author[\dagger]{Beidi Chen}

\affiliation[\dagger]{Carnegie Mellon University}
\affiliation[\ddagger]{Nvidia}

\abstract{
 Autoregressive Large Language Models (AR-LLMs) frequently exhibit implicit parallelism in sequential generation. Inspired by this, we introduce \textbf{Multiverse}, a new generative model that enables natively parallel generation. Multiverse internalizes a MapReduce paradigm, generating automatically through three stages: (i) a \texttt{Map} stage for adaptive task decomposition, (ii) a \texttt{Process} stage for parallel subtask execution, and (iii) a \texttt{Reduce} stage for lossless result synthesis. Next, we build a real-world Multiverse reasoning model with co-design of data, algorithm, and system, enabling rapid and seamless transfer from frontier AR-LLMs. For data creation, we develop \textit{Multiverse Curator}, an automated LLM-assisted pipeline that transforms sequential reasoning chains into structured training data, avoiding costly human annotations. Algorithmically, we design \textit{Multiverse Attention} to separate parallel reasoning steps while keeping compatibility with causal attention for efficient training. Systematically, we implement \textit{Multiverse Engine} to support parallel inference. It features a dedicated interpreter that dynamically switches between sequential and parallel generation, triggered directly by the model. After a 3-hour fine-tuning with 1K examples, \textit{our Multiverse-32B stands as the only open-sourced non-AR model achieving performance on par with leading AR-LLMs of the same scale}, evidenced by AIME24 \& 25 scores of 54\% and 46\%, respectively. Moreover, our budget control experiments show that Multiverse-32B exhibits superior scaling, outperforming AR-LLMs by 1.87\% on average using the same context length. Such scaling further leads to practical efficiency gain, achieving up to 2$\times$ speedup across varying batch sizes. We have open-sourced the entire Multiverse ecosystem, including data, model weights, engine, supporting tools, as well as complete data curation prompts and detailed training and evaluation recipes.
}

\metadata[Github]{\url{https://github.com/Multiverse4FM/Multiverse}}
\metadata[Website]{\url{https://Multiverse4FM.github.io}}

\begin{document}

\maketitle

\input{figures/introduction/intro}

\input{text/introduction}
\input{text/related}

\input{text/observation}
\input{text/methodology}

\input{text/instantiation}
\input{text/experiments}
\input{text/analysis}
\input{text/conclusion}

\clearpage
\newpage
\bibliographystyle{assets/plainnat}
\bibliography{paper}

\clearpage
\newpage
\beginappendix
\input{text/appendix}

\end{document}

%% file: figures/introduction/intro.tex
\begin{figure}[ht]    
    \centering 
    \includegraphics[width=\textwidth]{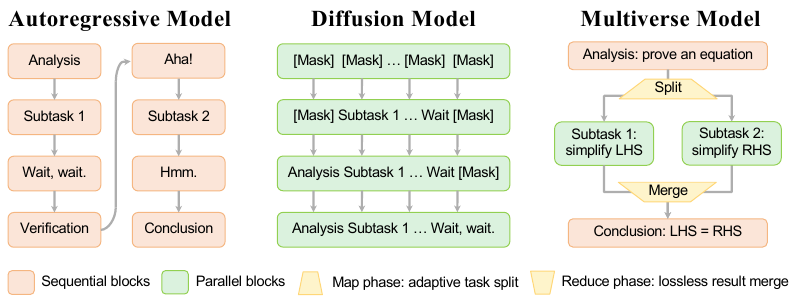}
        \caption{\textbf{Model Overview.} Autoregressive models are limited to sequential \mbox{generation, while diffusion models ignore} logical dependency in parallel generation. In contrast, \textbf{Multiverse} models enable adaptive and lossless parallel generation.}
    \label{fig:intro}
\end{figure}

%% file: text/introduction.tex
\section{Introduction}
\label{sec:introduction}

\begin{center}
    \textit{``In an infinite multiverse, everything that can happen does happen—somewhere."}
\end{center}

Test-time scaling has advanced Large Language Models (LLMs) by increasing the generation length~\citep{jaech2024openai, guo2025deepseek} and depth~\citep{geiping2025scaling}, closely reflecting human cognition. However, empowered by modern hardware like GPUs, ideal LLMs can surpass humans by \mbox{scaling a third dimension: \textit{width}}, which allows parallel task-solving. Realizing this potential requires LLMs to ``smartly'' parallelize and merge their generation, following \mbox{the classic \textit{MapReduce} paradigm~\citep{dean2004mapreduce}: splitting into subtasks, processing} them independently in parallel, and merging their results. Such philosophy has a long history in computer science~\citep{mccarthy1960recursive, aho1974design} while driving fundamental progress in other fields including  manufacturing~\citep{hounshell1984american}, agriculture~\citep{netting1993smallholders}, and finance~\citep{dean2008mapreduce}. This shift from sequential to parallel task-solving unlocks economies of scale: \mbox{reducing the time per unit and keeping near-constant} overall latency as task complexity grows, thereby offering a promising path towards artificial superintelligence (ASI).

Despite this potential, current LLMs are limited by the inherently sequential nature of autoregressive (AR) generation. While non-AR architectures, such as diffusion models~\citep{sahoo2024simple, zhao2025d1} and consistency models~\citep{kou2024cllms}, along with their hybrid semi-AR variants~\citep{arriola2025block, nie2025large}, natively enable parallel generation, they incur substantial computational waste. Their rigid, brute-force parallelism ignores logical dependencies, partly due to a lack of real-world training data to supervise when and how parallel generation should occur. Another stream of research~\citep{zhang2024accessing, cobbe2021training, yao2023tree, pan2025learning} leverages external tools to parallelize or merge tasks heuristically, leading to the loss of internal states, like the intermediate reasoning steps, \mbox{during communication with external modules. Although our concurrent} work~\citep{jin2025learning, rodionov2025hogwild} allows internal communication, they introduce inconsistencies between training and inference, limiting their effectiveness to short sequences with shallow parallelism. These challenges raise a research question: \textit{How to design a modeling framework for LLMs that can (i) adaptively split and merge tasks, (ii) losslessly preserve internal states, and (iii) generally apply to diverse parallelism patterns?}

Due to the dominance of AR-LLMs, we start to answer it by revealing \textit{\mbox{numerous intrinsic parallelism} in the sequential outputs of these models}. Specifically, we analyze the long Chain-of-Thought (CoT) trajectories from the s1K-1.1 dataset~\citep{muennighoff2025s1}. Among them, over 98\% exhibit parallelizable branches, despite being trained only for sequential generation.
These branches, as shown in Figure~\ref{fig:observation}, fall into collective and selective ones that appear frequently within individual CoT trajectories, either consecutively or recursively, covering a wide range of scenarios.
However, our prompting and probing tests verify that AR-LLMs cannot actively enforce or discern such parallelism. These findings motivate the design of a new modeling framework that can be bootstrapped directly from pre-trained AR-LLMs, which further requires us to address three practical limitations: (i) \textit{Data}: Real-world CoT trajectories lack explicit parallelism. (ii) \textit{Algorithm}: Transformers with causal attention are limited to sequential \mbox{generation. (iii) \textit{System}: Inference engines for AR-LLMs cannot execute parallel generation.}

To achieve these, we introduce \textbf{Multiverse}, a generative modeling framework built on the MapReduce paradigm that dynamically adjusts its parallelism during generation. It internalizes a three-stage pipeline: a sequential \texttt{Map} stage performs adaptive task decomposition; a parallel \texttt{Process} stage allows independent subtask execution; and a sequential \texttt{Reduce} stage ensures lossless result synthesis. Moreover, the pipeline can invoke itself recursively, enabling optimal time complexity with unlimited resources. We theoretically prove this optimality on a synthetic NP-hard SAT problem, demonstrating that Multiverse is the only framework that achieves a linear-time solution. Building on this, we co-design our data, algorithm, and system, providing a universal approach to building a real-world Multiverse model for complex reasoning \mbox{tasks, offering a seamless and rapid transition from AR-LLMs.} 

\textbf{Data Curation.} In Section~\ref{sec:p1-data}, we develop \mbox{\textbf{Multiverse Curator},  an automated LLM-assisted pipeline that} transforms sequential reasoning chains into parallel structures via five steps: (i) parsing the sequential chain into a summary tree; (ii) identifying parallelizable nodes within the summary tree; (iii) reformatting the summary into a parallel generation structure; (iv) refilling original reasoning steps into this structure; and (v) adding \texttt{Map} \& \texttt{Reduce} stages while rewriting \texttt{Process} stage. Moreover, content and grammar checks are performed to flag low-quality data for regeneration, avoiding costly manual filtration and annotation. In practice, this process results in \textbf{Multiverse-1K}, a dataset of \mbox{1,000 high-quality structured training samples for advancing LLM reasoning.}

\textbf{Algorithm Design.} In Section~\ref{sec:p1-model}, we design \textbf{Multiverse Attention} to enable parallel generation while keeping training efficiency. This is achieved by modifying attention masks and position embeddings to strictly separate independent reasoning branches in attention calculation, which can be trained in parallel, similar to causal attention. This design also excels in data efficiency: since these changes are minor, pre-trained AR models can be rapidly transferred from \mbox{causal attention to Multiverse attention using only a few thousand examples.}

\textbf{System Implementation.}  In Section~\ref{sec:p1-system}, we implement \textbf{Multiverse Engine} featuring a specialized interpreter to support MapReduce execution. By interpreting control tags generated by Multiverse models, our engine can dynamically switch between sequential and parallel generation without overhead, yielding a flexible workflow. This includes (i) Sequential $\rightarrow$ Parallel: mapping subtasks to separate branches for parallel execution with prefix sharing, and \mbox{(ii) Parallel $\rightarrow$ Sequential: reducing Key-Value (KV) states from all branches back into one sequence.}

The integration of these modules enables efficient training and inference of Multiverse models. Specifically, we develop Multiverse-32B by applying supervised fine-tuning (SFT) to Qwen-2.5-32B-Instruct using only 1K examples within 3 hours. Empirically, Multiverse-32B achieves significant improvements in reasoning ability, outperforming the base model by 23.6\%, with AIME24 and AIME25 scores of 53.8\% and 45.8\%, respectively. These results are comparable to  AR-LLMs, confirming that Multiverse does not compromise model performance. Furthermore, Multiverse-32B exhibits more efficient test-time scaling, yielding an average improvement of 1.87\% within fixed latency constraints. This efficiency stems from its parallel generation capabilities, leading to up to 2$\times$ wall-clock speedup per generated token \mbox{while maintaining effective scaling across variable batch sizes range from 1 to 128.}

We have open-sourced the entire Multiverse ecosystem, including data, model weights, engine, and supporting tools, along with complete data curation prompts and detailed training and evaluation recipes. We hope this full-stack release will inspire and accelerate advancements in developing more \mbox{efficient and scalable generative models.}

%% file: text/related.tex
\section{Related Work}
\label{related_work}

\textbf{Test-time Scaling.} Prior work has shown that optimizing AR-LLMs to generate longer outputs improves their reasoning abilities. This is evident in frontier reasoning models built with reinforcement learning (RL)~\citep{openai2024openaio1card, deepseekai2025deepseekr1incentivizingreasoningcapability, openai-o3, google-gemini-2.5, grok-3}, and also validated through supervised fine-tuning (SFT) on smaller models with a few distilled examples~\citep{muennighoff2025s1, ye2025limo}. However, this length scaling greatly increases latency due to the sequential nature of AR generation. Other methods like depth scaling~\citep{geiping2025scaling, zhao2025d1} suffer from the same issue, while width scaling~\citep{brown2024large, pan2025learning} requires external tools/models to split or merge generations.

\textbf{Internal Parallel Generation.} Recent work has increasingly explored other models to replace the commonly used AR models, thereby enabling parallel generation. Among them, discrete diffusion models~\citep{sahoo2024simple, shi2024simplified, austin2021structured, lou2023discrete, wang2025remasking}, including masked and absorbed variants, are gaining growing attention. To narrow their gap with AR models, efforts have been made on methods like hybrid AR-diffusion generation~\citep{arriola2025block, fathi2025unifying} and training/test-time scaling~\citep{nie2025large, zhao2025d1, dream2025}. However, \citep{feng2025theoretical} has theoretically shown that these approaches cannot reduce the number of sequential generating or sampling steps, as they brute-force parallelize token generation without adhering to inherent relations. Similarly, other work explores continuous diffusion models~\citep{barrault2024large} and consistency models~\citep{kou2024cllms}. Among these open-sourced, non-AR models, a common issue is their current inability to scale to complex reasoning tasks, such as AIME~\citep{aime2024}. While our concurrent work~\citep{jin2025learning, rodionov2025hogwild} begins to explore the use of customized attention masks for parallel generation, their design are not general or adaptive, limiting their effectiveness to shallow, non-nested parallelism. In contrast, our Multiverse offers a more efficient and scalable approach to enable \mbox{internal parallel generation, which is generally applicable to diverse parallelism patterns.}

\textbf{External Parallel Generation.} In another line of research, several approaches leverage external tools or models to enable parallel generation~\citep{yao2023tree, pan2025learning, wang2022self, brown2024large, zhang2024accessing}. However, these methods generally leverage heuristic rules and external tools to parallelize or merge their generation. For instance, Best-of-N~\citep{brown2024large} and self-consistency~\citep{wang2022self} use a brute-force approach by parallelizing generation at the beginning of generation. Other methods like Monte Carlo tree search (MCTS)~\citep{zhang2024accessing} and Tree of Thoughts (ToT)~\citep{yao2023tree} offer more fine-grained parallelism, yet they are still fundamentally guided by heuristics and depend on an external verifier. While recent work~\citep{pan2025learning} enables more adaptive parallel generation, it suffers from significant information loss when parallelizing and merging branches, as it requires inter-model communication when switching between sequential and parallel generation, during which short \mbox{text summaries rather than complete KV states can be shared.}

%% file: text/observation.tex
\section{Long CoT Generation: Sequential or Parallel in  Logic?}
\label{sec:observation}

This section presents several key observations of parallelism potential in AR-LLMs. First, in Section~\ref{sec:3.1}, we examine the CoT outputs from such models, verifying the common existence of intrinsic parallelism. Next, Section~\ref{sec:3.2} details two tests, showing that AR-LLMs \mbox{cannot explicitly enforce or discern this parallelism during generation.}

\subsection{LLMs can Implicitly Generate Parallelizable Branches.}
\label{sec:3.1}

We start by analyzing the long CoT trajectories of AR-LLMs using the s1K-1.1 dataset~\citep{muennighoff2025s1}, including both Deepseek R1~\citep{guo2025deepseek} and Gemini 2.0 Flash Thinking~\citep{Google:GeminiThinkingMode2024}, aiming to answer the \mbox{research question: \textit{Does the logic of sequentially generated tokens truly rely on all context that precedes it? }}

\input{figures/observation/observation}

Surprisingly, we reveal instances of  ``parallelizable branches", where multiple independent logic steps can execute concurrently, rather than strictly sequentially. These branches highlight the inherent parallelism in AR-LLMs. Figure~\ref{fig:observation} classifies them as collective or selective, which appearing flexibly in consecutive and recursive structures.

\textbf{Collective Branches} involve splitting a task into multiple subtasks can be processed concurrently, whose outputs are merged into a final result. Examples \mbox{include studying various cases and analyzing individual problems.}

\textbf{Selective Branches} refer to scenarios where numerous potential paths are considered, but not all necessarily contribute to the final output. Examples include exploring diverse solutions or examining competing hypotheses.

\begin{table}[htbp]
  \centering
  \caption{\textbf{Statistics of Implicit Parallelism.} Parallelizable branches commonly exist in long CoT trajectories generated by AR-LLMs. Per-example \mbox{existence ratio ($\mathrm{R}$\%) and frequency ($\mathrm{F}$) of different types are measured in the format $\mathrm{R|F}$.}} 
  \resizebox{\linewidth}{!}{\setlength{\tabcolsep}{1mm}{
  \begin{tabular}{lcccccc}
    \toprule
    & \multicolumn{3}{c}{\textbf{Collective Branch}} & \multicolumn{2}{c}{\textbf{Selective Branch}} & \multicolumn{1}{c}{\multirow{2}{*}{\textbf{Total}}} \\
    \cmidrule(lr){2-4} \cmidrule(lr){5-6} 
    & Case Study & Subtask Execution & Other & Path Exploration & Other & \\
    \midrule
    Deepseek R1    &   28.0 $|$ 2.00     &  47.3 $|$ 3.38    &  3.2 $|$ 0.23  &  19.4 $|$ 1.39   & 1.1 $|$  0.07          &  99.0 $|$  7.07  \\
    Gemini 2.0 Flash  & 39.4 $|$ 2.82   & 45.0 $|$ 3.22  &    2.9 $|$ 0.21  &   \hspace{0.5em}8.3 $|$ 0.60   &   1.4 $|$ 0.09     &   97.0 $|$ 6.94   \\
    \bottomrule
  \end{tabular}}}
  \label{tab:obs1} 
\end{table}

Table~\ref{tab:obs1} further details the occurrence ratios and frequencies for various types within the s1K-1.1 data. Notably, over 98\% of examples feature parallelizable branches. Among them, collective branches, such as case studies and subtask executions, are predominant, accounting for 79\%, with selective branches like path exploration comprising the remaining 19\%. Moreover, these \mbox{branches occur frequently, appearing an average of 7 times each example.}

\subsection{LLMs cannot Explicitly Structure Parallelizable Branches.}
\label{sec:3.2}
Next,we show that AR-LLMs cannot explicitly enforce or discern such parallelism from token and hidden spaces.

\begin{figure}[htbp]
  \centering
  \begin{subfigure}{0.67\textwidth}
    \centering
    \includegraphics[width=\textwidth]{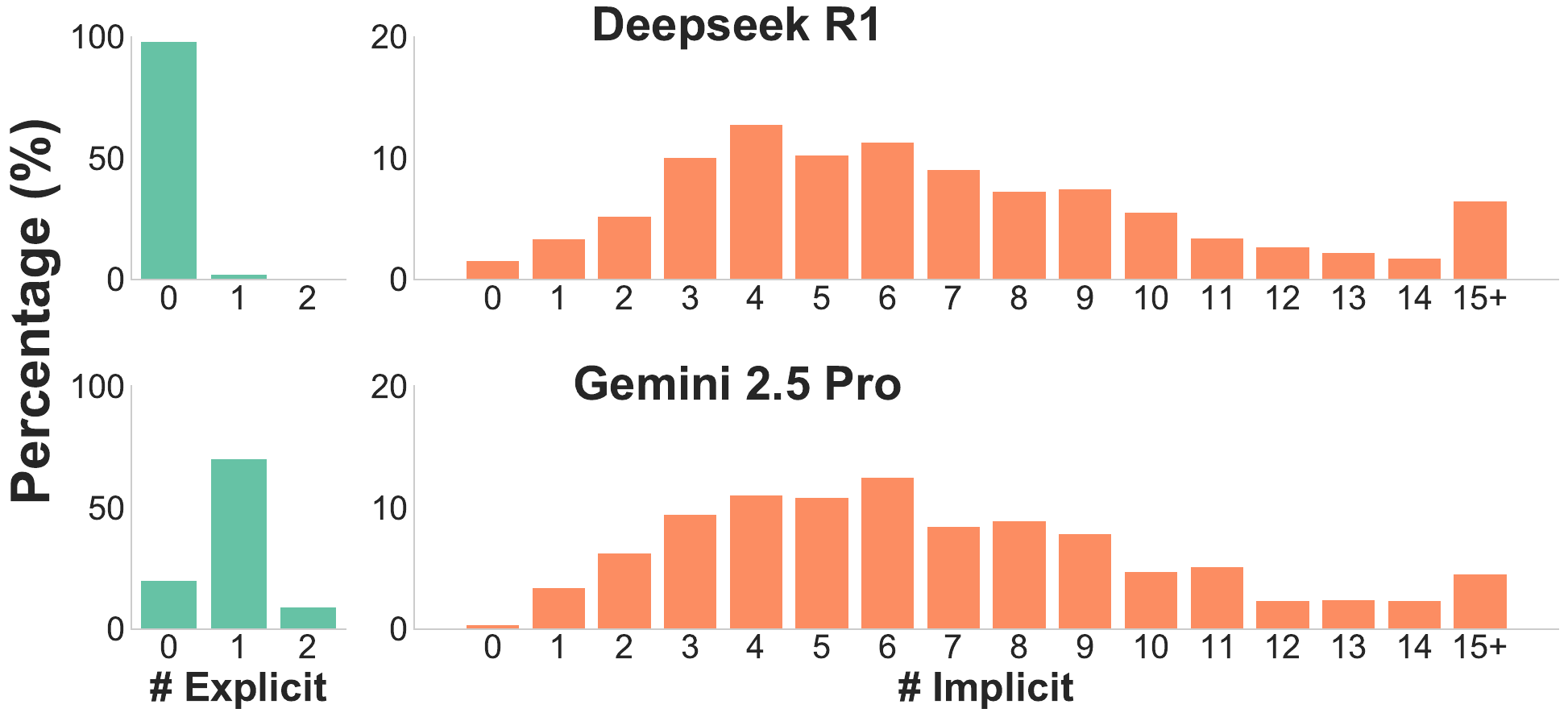}
    \caption{Prompting Test: Comparing Explicit and  Implicit Structure Counts}
    \label{fig:observation_explicit_vs_implicit}
  \end{subfigure}
  \hfill
  \begin{subfigure}{0.31\textwidth}
    \centering
    \includegraphics[width=\textwidth]{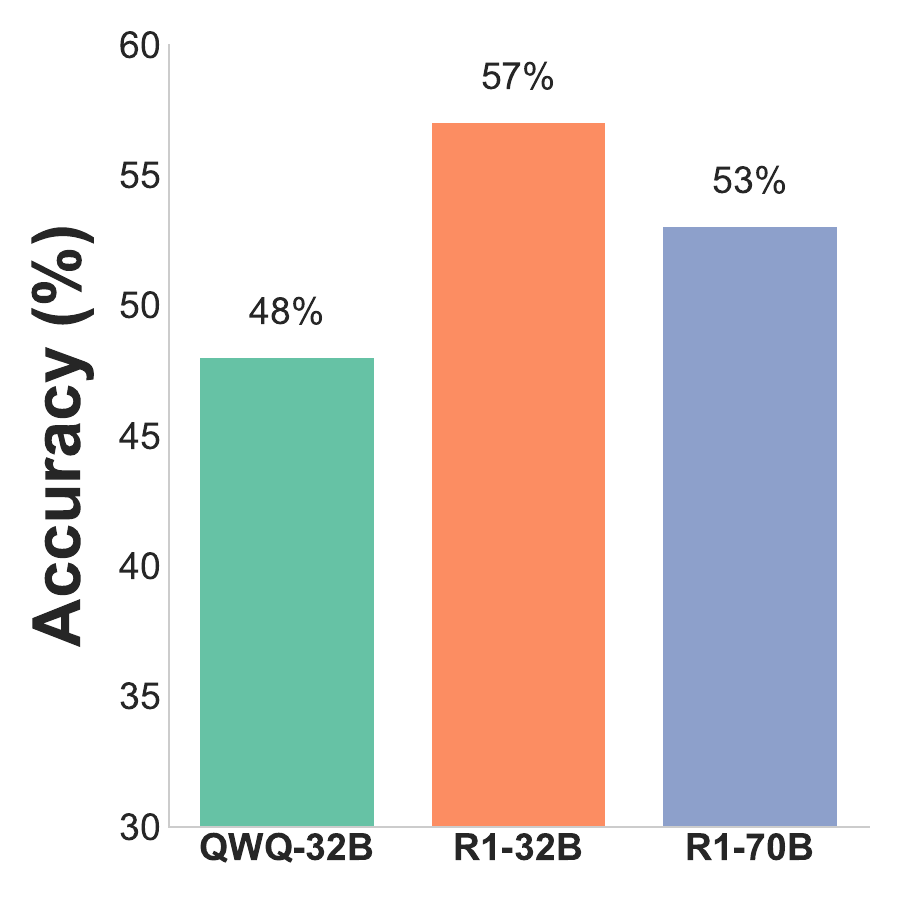}
    \caption{Probing Test: Classifier Accuracy}
    \label{fig:probing}
  \end{subfigure}

  \caption{Our prompting and probing tests show AR-LLMs cannot explicitly \textit{enforce or discern} parallelizable branches.}
  \label{fig:observation2}
\end{figure}

\textbf{Prompting Test.} We first prompt Deepseek R1 and Gemini 2.5 Pro using the same questions, with a detailed description of all valid parallel structures. Figure~\ref{fig:observation_explicit_vs_implicit} showcases a significant 90\% disparity emerges between explicit occurrence and implicit existence of \mbox{these structures, indicating that AR-LLMs struggle to generate in parallel.}


\textbf{Probing Test.} We then probe into the hidden space of AR-LLMs, confirming whether they can discern intrinsic parallelism. Specifically, we label tokens before parallel blocks as positive examples and treat all others as negative. Final-layer representations of these tokens are extracted using DeepSeek-R1-Distill-Qwen-32B \& 70B~\citep{deepseekai2025deepseekr1incentivizingreasoningcapability} and QWQ-32B~\citep{qwq32b}. A two-layer MLP classifier is trained to predict whether a token initiates parallelizable branches. However, the classifier's performance, which was comparable to random guessing (as shown in Figure~\ref{fig:probing}), suggests that AR-LLMs do not truly understand such parallelism. Instead, they generate these structures unconsciously, based on patterns learned from their pre-training corpus.


%% file: figures/observation/observation.tex
\begin{figure}[ht]    
    \centering 
\includegraphics[width=\textwidth]{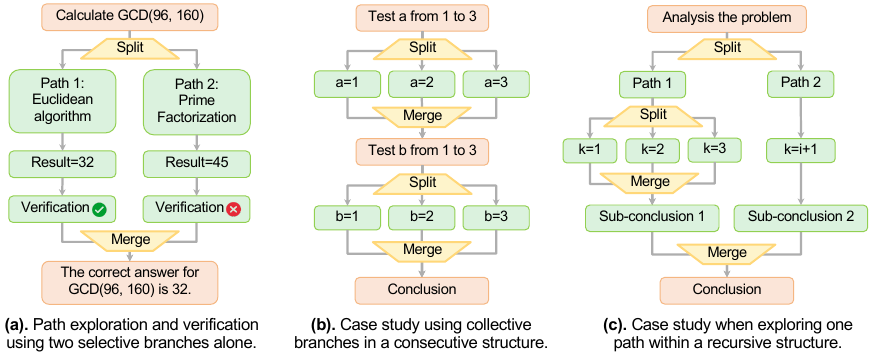}
        \caption{\textbf{Existence of Implicit Parallelism.} Parallelizable branches fall into collective (with all branches contributing) and selective (with some branches contributing) ones. They occur alone or as a part in consecutive or recursive structures.}
    \label{fig:observation}
\end{figure}

%% file: text/methodology.tex
\section{Designing Multiverse for Natively Parallel Generative Modeling.}
\label{sec:model}

With all findings in Section~\ref{sec:observation}, we present Multiverse, a novel generative modeling framework built on the MapReduce paradigm, which adaptively parallelizes and losslessly merges its generation to surpass AR models.


\subsection{Preliminaries.}

\textbf{Language Modeling} aims to learn the joint probability distribution $P(x^1, x^2, \ldots, x^L)$ over sequences of words or tokens given a finite vocabulary $V$ of tokens, and a sequence of $L$ tokens denoted by $\mathbf{x}^{1:L} = (x^1, x^2, \ldots, x^L)$.

\textbf{Autoregressive Modeling} represents sequence $\mathbf{x}^{1:L}$ from left to right, where each token $x_t$ is conditioned on all past tokens $\mathbf{x}^{1:t-1}$. Thus, the joint probability of $\mathbf{x}^{1:L}$ is factorized as a product of conditional probabilities:

$$P(\mathbf{x}^{1:L} |\theta_{AR}) = P(x^1, x^2, \ldots, x^L |\theta_{AR}) = \prod_{t=1}^{L} P(x_t  |x_1, \ldots, x_{t-1}; \theta_{AR})$$

where $\theta_{AR}$ denotes model parameters. AR models offer high accuracy but exhibit poor parallelism in generation.

\subsection{Multiverse Modeling.}

\input{figures/model/model}

Our modeling framework, Multiverse, advances beyond AR by eliminating redundant sequential dependencies between independent contexts, allowing for adaptive and lossless control over the start and end of parallel generation. To realize this, we adopt a MapReduce structure internalizing three stages, as illustrated in Figure~\ref{fig:model}.

\textbf{Map Stage.} The pipeline begins by sequentially generating a concise task decomposition plan, denoted as $\mathbf{x}_s$. Each subtask in $\mathbf{x}_s$ is then mapped to an independent prefix sequence, modeled as $P(\mathbf{x}_{1,s} |\mathbf{x}_s)$ and $P(\mathbf{x}_{2,s} | \mathbf{x}_s)$.

\textbf{Process Stage.} Next, it performs parallel modeling for each branch independently, conditioned on its own prefix. This enables the concurrent generation of diverse branches, like: $P(\mathbf{x}_{1,1:6} | \mathbf{x}_{[1:3,s]},\mathbf{x}_{1,s} )$ and $P(\mathbf{x}_{2,1:6}| \mathbf{x}_{2,s},\mathbf{x}_{[1:3,s]})$. Each branch ends when a specific suffix (i.e, $\mathbf{x}_{1,e}$ or $\mathbf{x}_{2,e}$) is generated. We use the same suffix for all branches.

\textbf{Reduce Stage.} After completing all branches, Multiverse shift back to sequential generation to conclude them, which is conditioned on preceding tokens from all branches, modeled as $P(\mathbf{x}_{e, [3:4]} | \mathbf{x}_{1,[s,1:6,e]}, \mathbf{x}_{2,[s,1:6,e]}, \mathbf{x}_{[1:3,s]})$.

This structure \mbox{enables Multiverse to: (i) adaptively decide how to parallelize generation during the \texttt{Map} stage;} and (ii) retain information by ensuring every branch remains fully accessible in the \texttt{Reduce} stage and beyond. Notably, Multiverse generalizes to \mbox{both recursive and consecutive compositions of multiple MapReduce structures.}

\subsection{Structured Generation Flow.} 
\label{sec:generation_flow}

\begin{wrapfigure}{r}{0.4\textwidth}
  \centering
  \vspace{-15pt}  
  \includegraphics[width=0.38\textwidth]{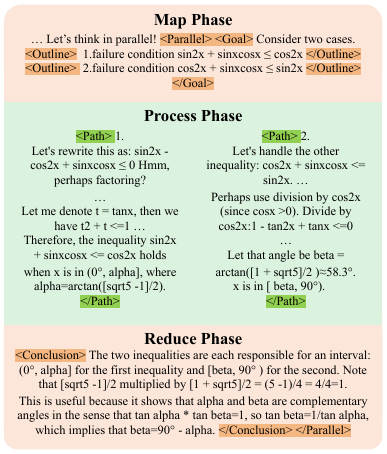}
  \vspace{-3pt}
  \caption{Example of MapReduce Structure.}
  \vspace{-30pt}
  \label{fig:structure}
\end{wrapfigure}

To enable automatic control over the generation flow, Multiverse further employs a structured set of specialized control tags that explicitly define each MapReduce block. These tags, such as \texttt{<Parallel>} and \texttt{<Path>}, delineate the boundaries of MapReduce blocks and coordinate the execution of all three internal stages. Figure~\ref{fig:structure} provides an example of this structure.

The MapReduce block begins with the \texttt{<Parallel>} tag, initiating the three-stage process. Immediately after it, the \textbf{Map} stage starts with the \texttt{<Goal>} tag, which defines the overall objective, which is then broken down into subtasks using multiple nested and indexed \texttt{<Outline>} tags. Following goal specification (signaled by \texttt{</Goal>}), the \textbf{Process} stage commences. At this stage, each subtask is independently mapped and processed within a \texttt{<Path>} block in parallel, matched by its index. Once all paths have finished (signaled by \texttt{</Path>}), the \texttt{<Conclusion>} tag triggers the \textbf{Reduce} stage that synthesizes the results from these independent paths into a final coherent output, which is ended with the \texttt{</Conclusion>} tag. Finally, the MapReduce block is terminated by the \texttt{</Parallel>} tag.

%% file: figures/model/model.tex
\begin{figure}[ht]   
    \centering 
\includegraphics[width=\textwidth]{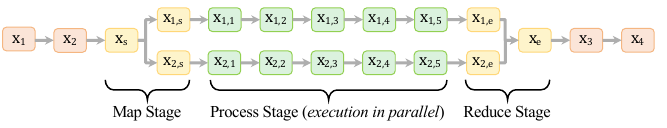}
        \caption{\textbf{Multiverse} enables adaptive and lossless parallel generation by natively internalizing a MapReduce paradigm.}
    \label{fig:model}
\end{figure}

%% file: text/instantiation.tex
\section{Building a Real-world Multiverse Model.}
\label{sec:method}


To deploy Multiverse in real-world scenarios, we present a comprehensive suite consisting of \textit{Multiverse Curator} as the data generator, \textit{Multiverse Attention} as the core algorithm, and \textit{Multiverse Engine} as the optimized system. This suite enables a smooth and rapid shift from leading AR models to Multiverse models. In particular, we apply this suite in complex reasoning tasks, leading to a \mbox{Multiverse model that exhibits strong reasoning capabilities.}

\input{figures/instantiation/intro}

\subsection{Data Curation: Multiverse Curator.}
\label{sec:p1-data}
While the long CoT trajectories generated by AR-LLMs often inherently contain MapReduce structures, explicitly generating them is difficult, as detailed in Section~\ref{sec:observation}. To address this absence of MapReduce structures in existing data, we introduce \textbf{Multiverse Curator}, an LLM-assisted pipeline that automatically transforms sequential reasoning chains into parallel MapReduce structures. This convert is guided by a five-stage prompting protocol powered by Gemini 2.5 Pro~\citep{Google:Gemini2.5Pro2025} in Figure~\ref{data}. \mbox{The detailed prompts are available in Appendix~\ref{app:prompt}.}

\textbf{Generating a Summary Tree.}  First, we iteratively decompose and outline the original reasoning chain into a two-level tree-structured summary. In the first round, the entire reasoning chain is broken down into multiple steps. In the second round, each step is examined by the LLM, with complex steps being further split into substeps. Finally, every identified step or substep will be clearly \mbox{labeled and outlined with a concise descriptive summary.}

\textbf{Identifying Parallel Groups.} Second, we instruct the LLM to analyze the relationship between consecutive reasoning steps, identifying which steps or step groups can execute in parallel \mbox{without violating logical dependencies.}

\textbf{Reformating into Parallel Structures.} Third, the summary tree is transformed into a parallel structure using the grouping results from the previous step. To signal parallel execution, parallelizable steps or step groups are explicitly marked by enclosing them \mbox{in the control tags  \texttt{<Parallel>} and \texttt{</Parallel>}, forming a parallel block.}

\textbf{Refilling Original Details.} Fourth, we prompt the LLM to repopulate the detailed content for each step and substep while keeping the structures. The LLM will retrieve and copy the \mbox{related texts from original trajectories.}

\textbf{Adding MapReduce Structures \& Rewriting All Paths}. Finally, we further convert the parallel structures into the MapReduce structures defined in Section~\ref{sec:generation_flow}. For each parallel block, the LLM generates both the \texttt{Map} and \texttt{Reduce} stages itself by outlining the specific targets and results for each individual path. Moreover, all paths are rewritten to avoid words implying sequential relations (e.g., ``Similarly" or ``Alternatively'') and to prevent including or referencing content \mbox{from other paths, ensuring the completeness and independence of each path.}

To further enhance our data quality, two supplementary validation stages have been incorporated. After the fourth stage, a content check will filter out data if its edit distance ratio is above 0.2. Next, after the fifth stage, a grammar check will confirm strict adherence to our MapReduce structures. Data failing either case will be iteratively regenerated through our pipeline until both standards are met. We provide more details in Appendix~\ref{app:prompt}. The application of this automated pipeline to the s1K-1.1 dataset has yielded \textbf{Multiverse-1K}, a new dataset consisting of 1,000 high-quality, \mbox{structured reasoning trajectories across a range of math and science problems.} 

\subsection{Algorithm Design: Multiverse Attention.} 
\label{sec:p1-model}
Next, we introduce \textbf{Multiverse Attention} to replace the causal attention~\citep{vaswani2017attention} in AR-LLMs. Causal attention computes the $i$-th token's output with query \(\bm{q}_i\), and keys \(\bm{k}_j\), values \(\bm{v}_j\) from positions \(j \leq i\):
\begin{equation}
a_{ij} = \text{Softmax}\left((\bm{q}_i^\top \odot P(i)) \cdot (\bm{k}_j \odot P(j)) + M_{ij}\right),
\end{equation}
where \(M_{ij} =
\begin{cases}
0, & j \leq i \\
-\infty, & \text{otherwise}
\end{cases}\) is the causal mask, and \(P(i)\) is the positional embedding for the \(i\)-th position. 

However, this formulation poses challenges for conceptual parallel generation, as later paths depend on both (i) the key-value (KV) pairs and (ii) the positional indices produced by earlier paths. To address this, we modify both the attention masks and position indices following APE~\citep{yang2025ape}, as shown in Figure~\ref{algo}. In Multiverse Attention, each path within the same \texttt{Process} block starts from an identical position and executes independently without accessing others. During the \texttt{Reduce} stage, all paths converge to the same position, which is set to the maximum position reached by any path to \mbox{avoid negative relative distance, regardless of their variable lengths.}

Building on its similarity to causal attention, Multiverse Attention enables (i) \textbf{Hardware Efficiency:}it can preserve training parallelism, and (ii) \textbf{Data Efficiency:} it can be rapidly \mbox{adapted via fine-tuning on a few samples.}

\subsection{System Implementation: Multiverse Engine.}
\label{sec:p1-system}
To enable truly parallel generation in practical deployments, we introduce \textbf{Multiverse Engine}, an extension of existing inference engines designed for AR models. Specifically, we start from SGLang~\citep{zheng2023efficiently} due to its support
for continuous batching and radix attention. These features allow dynamic batch scheduling and flexible KV-cache reuse for Multiverse, two scenarios that frequently occur in the \texttt{Map} and \texttt{Reduce} stages.

The \texttt{Map} stage is automatically triggered when a $\texttt{<Parallel>}$ token is generated. Next, the interpreter counts the number of $\texttt{<Outline>}$ encountered until reaching $\texttt{</Goal>}$. Based on this count, the engine creates multiple paths executed in parallel, which can be viewed as distinct samples within the same batch.  Leveraging radix attention, these paths share the prefix KV cache from the current context. Each path is identified and initiated with ``\texttt{<Path> i}" according to its order \texttt{i} in the \texttt{<Outline>} list. After prefilling, all paths are added to the decoding queue for parallel generation. When a path finishes, either by reaching \texttt{</Path>} or the maximum length, it enters a ``zombie" state that releases all resources and waits \mbox{for the completion of other paths before continuing to the next stage.}

The \texttt{Reduce} stage begins once all processing paths have completed their execution. During this stage, the engine merges the KV states from all paths along with the preceding context to form a new sequence. Thanks to the flexible memory layout of the radix cache, indices of KV cache can be seamlessly concatenated without any padding, thereby avoiding both physical memory copying overhead and redundant padding computations. A token \texttt{<Conclusion>}, prefixed with this merged KV cache, is subsequently added to the prefilling queue. Upon completion, the request advances \mbox{to the decoding queue to continue generation along the newly constructed sequence.}

%% file: figures/instantiation/intro.tex
\begin{figure}[H]   
    \centering 
    \begin{subfigure}[b]{\textwidth}
        \centering
        \includegraphics[width=\linewidth]{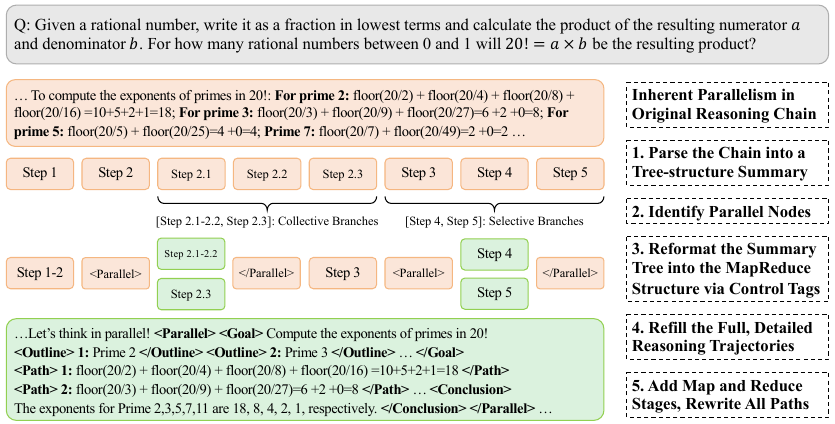} 
        \caption{\textbf{Multiverse Curator} automatically generated \textbf{Multiverse-1K} using an LLM-assisted data curation pipeline.}
        \label{data}
        \vspace{0.5em}
    \end{subfigure}

    \begin{subfigure}[b]{\textwidth}
        \begin{subfigure}[b]{0.44\textwidth} 
            \centering
            \includegraphics[width=\linewidth]{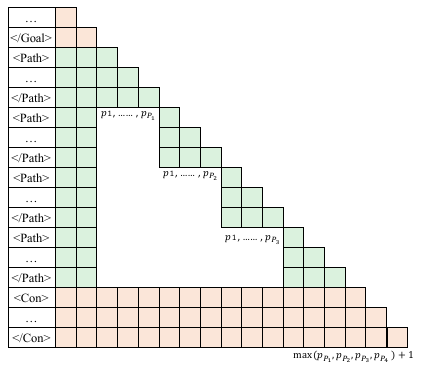} 
            \caption{\textbf{Multiverse Attention}} 
            \label{algo} 
        \end{subfigure}
        \hfill 
        \begin{subfigure}[b]{0.52\textwidth} 
            \centering
            \includegraphics[width=\linewidth]{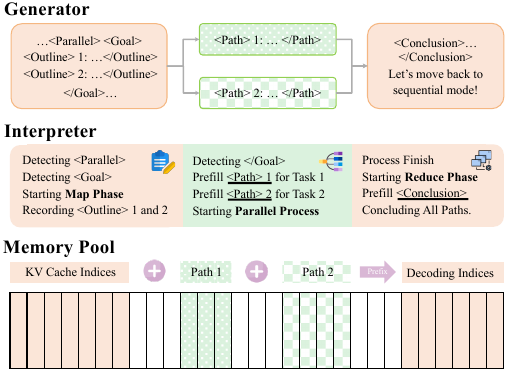} %
            \caption{\textbf{Multiverse Engine}} 
            \label{sys} 
        \end{subfigure}
    \end{subfigure}

    \caption{\textbf{Instantiation Overview.} Multiverse co-design data (\textit{Multiverse Curator}), algorithm (\textit{Multiverse Attention}), and system (\textit{Multiverse Engine}) to enable real-world reasoning abilities through a rapid and seamless shift from AR-LLMs.} 
    \label{fig:intro2}
\end{figure}

%% file: text/experiments.tex
\section{Experiments}
\label{sec:exp}
This section shows the superiority of Multiverse over Autoregression in real-world reasoning tasks. Specifically,
\begin{itemize}[itemsep=0.0ex,topsep=0pt,leftmargin=*]
    \item In Section~\ref{sec:performance}, Multiverse-32B achieves substantial improvements over the Qwen2.5 model by 24.5\% after SFT on Multiverse-1K, \mbox{while matching or exceeding the performance of AR-LLMs on real-world reasoning tasks.}
    \item In Section~\ref{sec:scaling}, Multiverse-32B exhibits a superior tradeoff between performance and latency than AR-LLMs. It achieves this by generating more tokens within the \mbox{same wall-clock time, indicating a more efficient scaling.}
\end{itemize}

\subsection{Setup.}
\textbf{Training.} We create Multiverse-32B by performing SFT on the Qwen2.5-32B-Instruct model~\citep{qwen2.5}, integrating our Multiverse Attention. The training dataset combines our Multiverse 1K dataset prompted with ``Think step by step and in parallel" and the original sequential data appended by ``Think step by step". We employ a dynamic mixture ratio that progressively shifts from 0:1 (exclusively Autoregressive data) to 1:0 (exclusively Multiverse data) across eight epochs. \mbox{Fine-tuning took 3 hours on 8 NVIDIA B200 GPUs with PyTorch FSDP.}

\textbf{Evaluation.} We measure Multiverse-32B on four reasoning tasks, including AIME24~\citep{aime2024}, AIME25~\citep{aime2025}, MATH500~\citep{hendrycks2021measuring}, and GPQA Diamond~\citep{rein2024gpqa}.  LightEval~\citep{lighteval} is employed as the evaluation toolkit, powered by our SGLang~\citep{zheng2023efficiently}-based Multiverse Engine. We evaluate our model under two prompting conditions: with and without the phrase ``in parallel'', where \mbox{the latter one is denoted as Multiverse-32B-zero}.

\textbf{Baselines.} We compare Multiverse-32B with the Qwen2.5 model and an Autoregressive-32B trained using the same data, but without any control tags or extra \texttt{Map} and \texttt{Reduce} stages. s1-32B and s1.1-32B~\citep{muennighoff2025s1} are also included for reference. In addition to pass@1, we measure the degree of parallelism (\# parallel) as the ratio between the number of generated tokens and the number of \textit{sequentially} generated tokens. 

\subsection{Real-world Reasoning Performance}
\label{sec:performance}

\begin{table}[h]
\centering
\caption{\textbf{Performance comparison between Multiverse-32B and other 32B AR-LLMs.} The pass@1 metric is reported using LightEval~\citep{lighteval}, with results averaging over 8 seeds on AIME. The \# parallel computes the ratio between the total number of generated tokens and the actual generation length, measuring the degree of parallelism.}
  \resizebox{\linewidth}{!}{\setlength{\tabcolsep}{1mm}{
\begin{tabular}{lcccccccc}
\toprule
 & \multicolumn{2}{c}{\textbf{AIME24}} & \multicolumn{2}{c}{\textbf{AIME25}} & \multicolumn{2}{c}{\textbf{MATH500}} & \multicolumn{2}{c}{\textbf{GPQA-Diamond}} \\
\cmidrule(lr){2-3} \cmidrule(lr){4-5} \cmidrule(lr){6-7} \cmidrule(lr){8-9}
\textbf{Model / Metric} & pass@1 & \# parallel & pass@1 & \# parallel & pass@1 & \# parallel & pass@1 & \# parallel \\
\midrule
\textcolor{darkgray!50}{s1-32B} & \textcolor{darkgray!50}{35.4} & \textcolor{darkgray!50}{1.00} & \textcolor{darkgray!50}{25.8} & \textcolor{darkgray!50}{1.00} & \textcolor{darkgray!50}{88.6} & \textcolor{darkgray!50}{1.00} & \textcolor{darkgray!50}{48.0} & \textcolor{darkgray!50}{1.00}\\
\textcolor{darkgray!50}{s1.1-32B} & \textcolor{darkgray!50}{52.9}& \textcolor{darkgray!50}{1.00} & \textcolor{darkgray!50}{41.7} & \textcolor{darkgray!50}{1.00} &  \textcolor{darkgray!50}{93.4}& \textcolor{darkgray!50}{1.00} & \textcolor{darkgray!50}{62.6} & \textcolor{darkgray!50}{1.00}\\
\cdashlinelr{0-8}
Qwen2.5-32B-Instruct & 15.8 & 1.00 & 10.4 & 1.00 & 80.4 & 1.00 & 47.0 & 1.00\\
Autoregressive-32B & \textbf{54.6} & 1.00 & \underline{45.0} & 1.00 & \textbf{92.8} & 1.00 & \underline{61.6} & 1.00 \\
\midrule
\textbf{Multiverse-32B-zero} & 52.1 & 1.04 & 44.2 & 1.05 &  \underline{92.4} & 1.12 & \textbf{63.6} & 1.17\\
\textbf{Multiverse-32B} & \underline{53.8} & 1.18 & \textbf{45.8} & 1.15 & 91.8 & 1.15 & 60.7 & 1.17\\
\bottomrule
\end{tabular}}}
\label{tab:main_exp}
\end{table}

In Table~\ref{tab:main_exp}, we report the performance of Multiverse-32B on complex reasoning tasks with 32K contexts, showing improvements of 38\%, 35\%, 11\%, and 14\% over the Qwen2.5-32B-Instruct model across the respective benchmarks after fine-tuning. Notably, Multiverse-32B matches or even surpasses the performance of autoregressive models, as demonstrated by its comparison with Autoregressive-32B. For reference, we also include the results of the s1.1-32B model trained on the sequential CoT data from which Multiverse-1K is derived. The comparable performance between these models confirms \mbox{that our data curation pipeline successfully preserves the original data quality.}

We also evaluate Multiverse-32B-Zero, a variant prompted without the “think in parallel” instruction. Comparing the two variants reveals distinct performance patterns: Multiverse-32B achieves greater parallelism on AIME tasks, resulting in a slight performance improvement, while Multiverse-32B-Zero performs better on tasks requiring shorter generation sequences, where the model naturally generates in parallel without explicit prompting. This parallelism, measured as the ratio of generated tokens to generation length, aligns with our training strategy, suggesting the potential for controllably switching between AR and Multiverse generation. Notably, the reduced parallelism observed on AIME tasks indicates that the model exhibits less parallelism during longer generation, which we attribute partly to the scarcity of \mbox{training data exceeding 16K tokens in Multiverse-1K.}

\subsection{Scaling Performance}
\label{sec:scaling}

\begin{figure}[h]
    \centering
    \begin{subfigure}[b]{0.49\textwidth}
        \includegraphics[width=\textwidth]{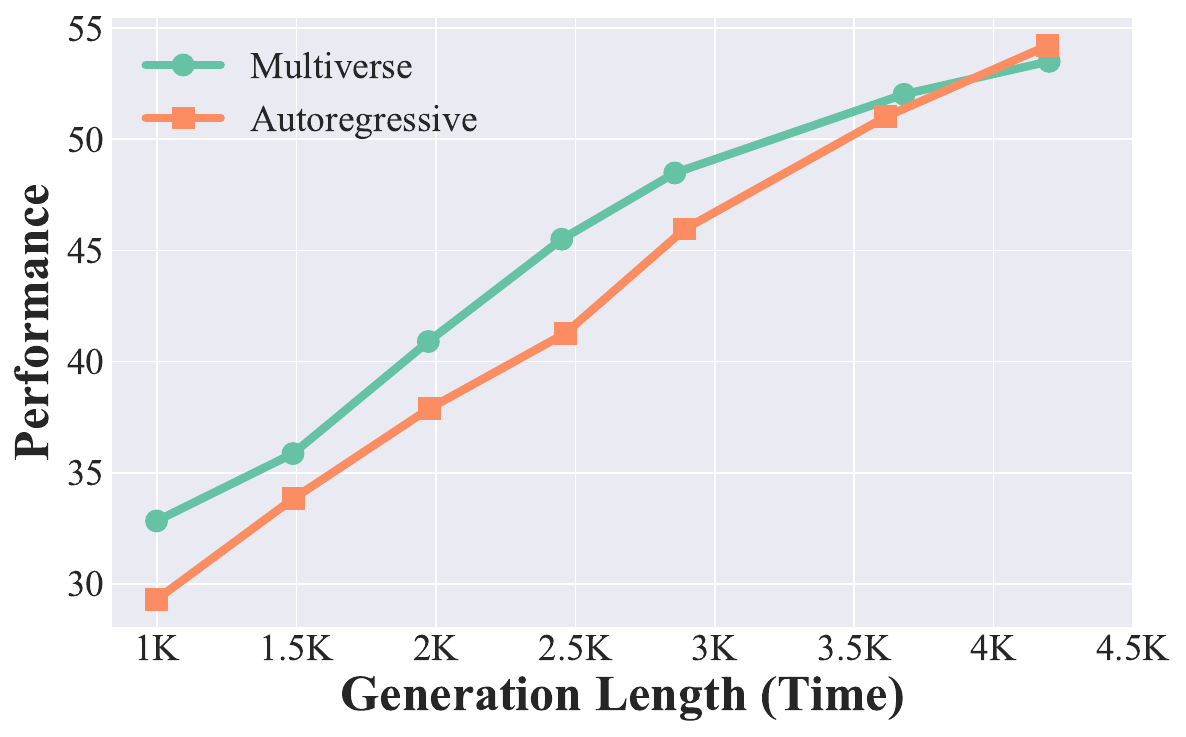}
        \caption{GPQA-Diamond}
        \label{fig:linear_scaling_1}
    \end{subfigure}
    \hfill
    \begin{subfigure}[b]{0.49\textwidth}
        \includegraphics[width=\textwidth]{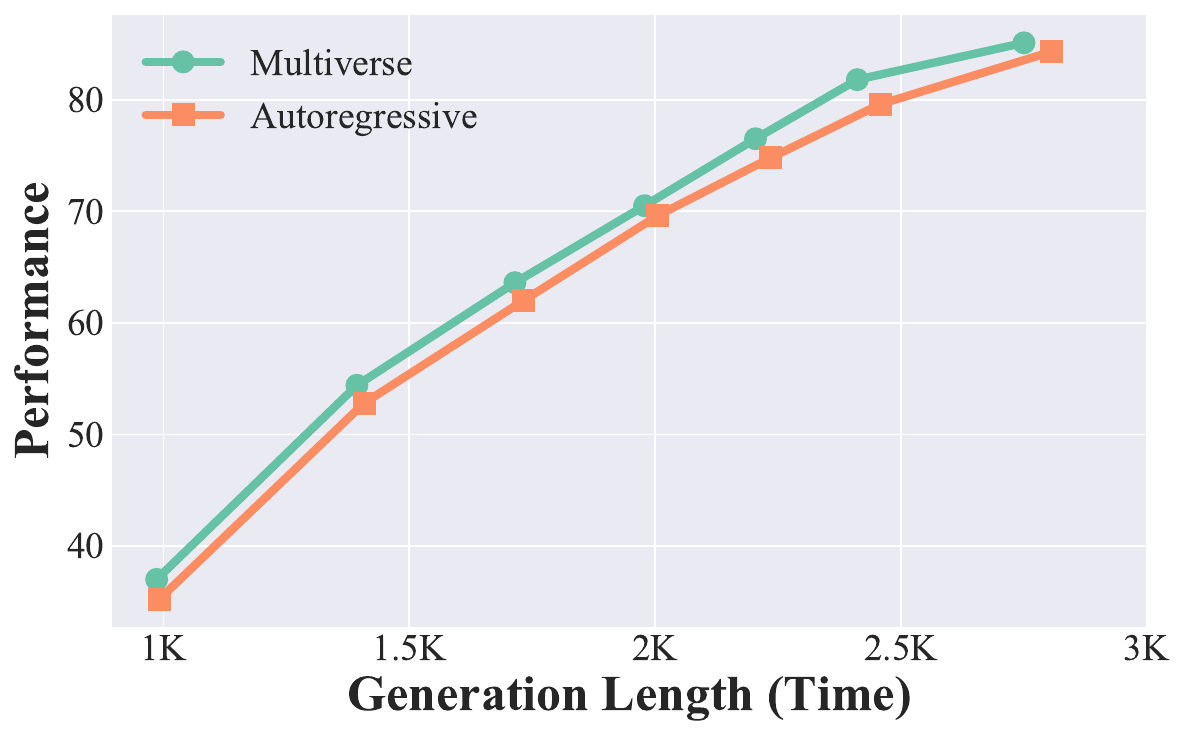}
        \caption{Math500}
        \label{fig:linear_scaling_2}
    \end{subfigure}
    \hfill

    \caption{Multiverse achieves better performance using a fixed context length, which indicates the same generation time. Here, we report the actual generation length, as some data points terminate before reaching the maximum length.}
    \label{fig:sequential_scaling}
\end{figure}

To highlight the benefits of parallel generation, we conduct budget control experiments on GPQA-Diamond and MATH500 using the same context length (i.e., approximately the same generation time),  varying from 1K to 4K tokens. As illustrated in Figure~\ref{fig:sequential_scaling}, while longer contexts improved performance for both models, Multiverse-32B generates more tokens within the same context length. This parallel scaling yielded performance improvements of 2.23\% on the GPQA-Diamond (with \# parallel = 1.17) \mbox{and 1.51\% on the MATH500 (with \# parallel = 1.15).}

%% file: text/analysis.tex
\section{Efficiency Analysis}
Having demonstrated Multiverse-32B's strong scalability and overall performance, we now further analyze the practical efficiency of Multiverse with our engine, showing the potential unlocked through parallel generation.

\begin{figure}[htbp]
    \centering
    \begin{subfigure}[b]{0.48\textwidth}
        \includegraphics[width=\textwidth]{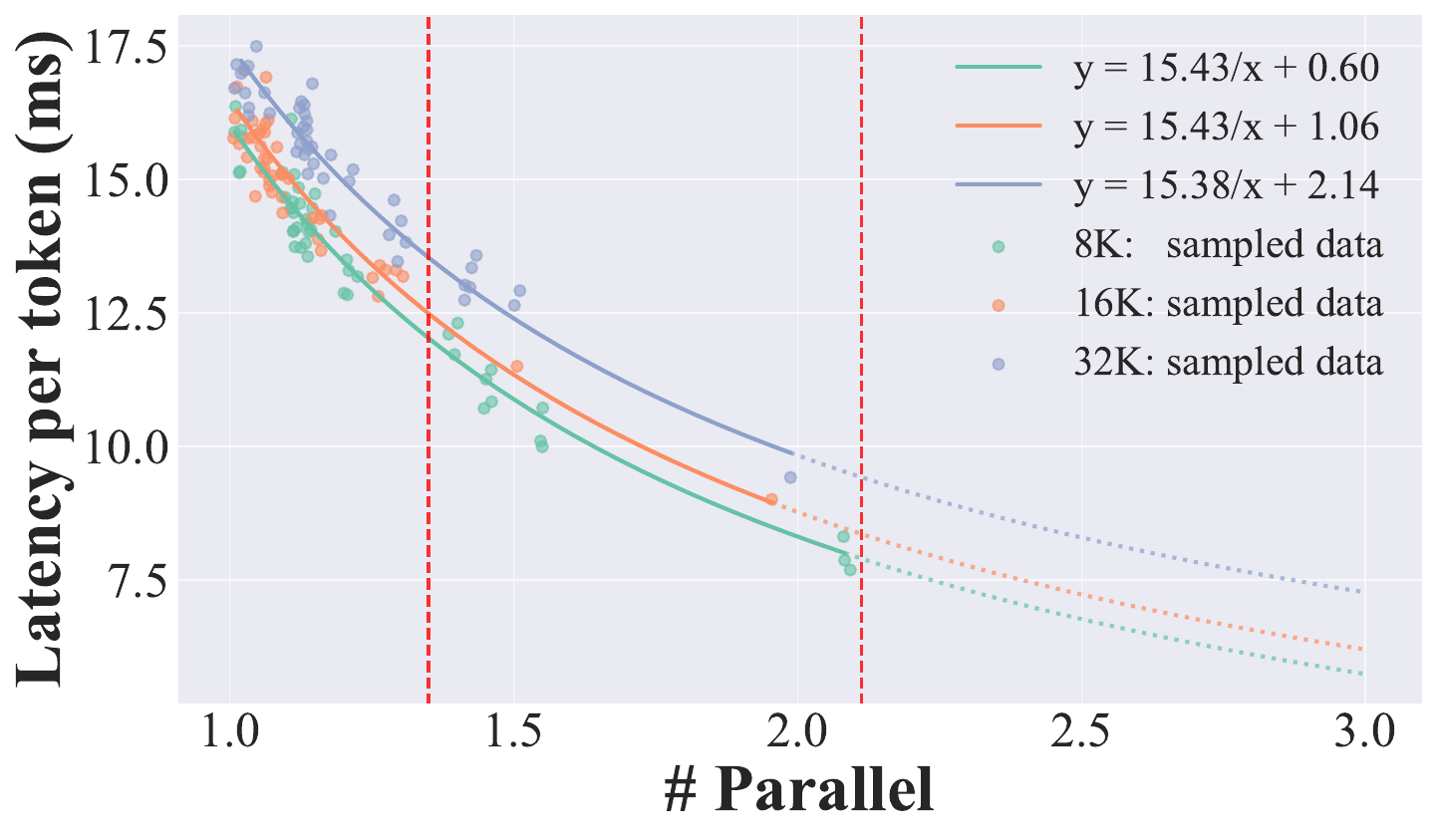}
        \caption{Reduced Latency/Token with Increased \# Parallel}
        \label{fig:latency}
    \end{subfigure}
    \hfill
    \begin{subfigure}[b]{0.48\textwidth}
        \includegraphics[width=\textwidth]{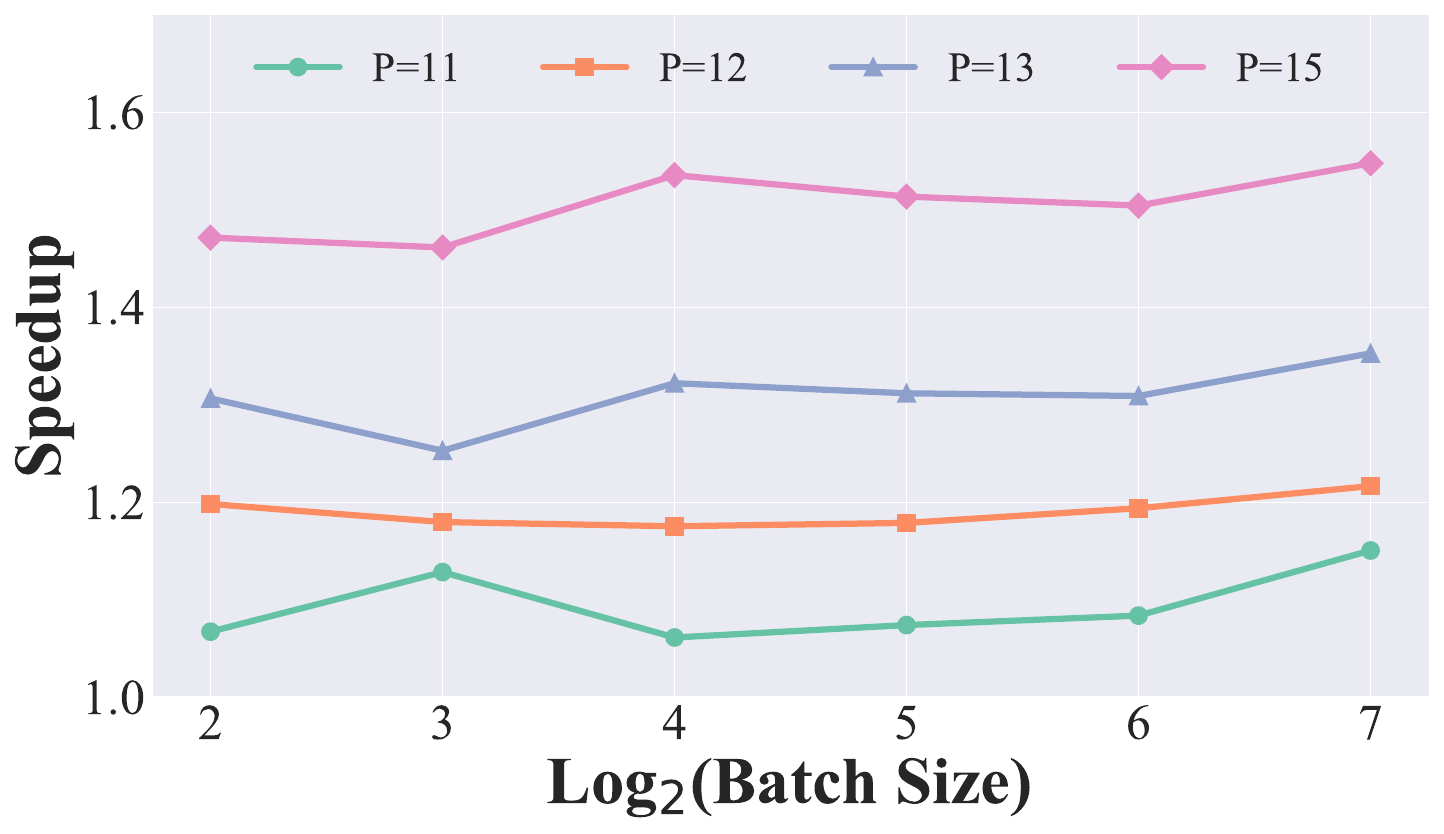}
        \caption{Stable Speedup Across Varying Batch Size}
        \label{fig:speedup_vs_bsz}
    \end{subfigure}

    \caption{\textbf{Efficiency Analysis.} Multiverse can obtain efficiency gains across varying batch sizes based on \# Parallel. }
    \label{fig:efficiency}
    \vspace{-1em}
\end{figure}

First, we investigate the relationship between the degree of parallelism and latency per token across various generation lengths (8K, 16K, and 32K), using a batch size of one. The resulting data points, illustrated in Figure~\ref{fig:latency}, demonstrate that Multiverse enhances generation efficiency by increasing the degree of parallelism. Furthermore, we fit the sampled data points into three inverse curves, one for each. These curves highlight the potential of Multiverse to further reduce latency by encouraging parallelism. Specifically, we identify three key regions based on the sample distribution, demarcated by red lines. The first, encompassing parallelism degrees from 1.0 to 1.3, represents the majority of data points, yielding an average speedup of 18.5\%. In the second region, examples show that higher parallelization is achievable, offering acceleration of up to 2.1$\times$. Finally, the third region, characterized by extended lines, demonstrates the \mbox{promising potential for further improvement with increased parallelism.}

Next, we show the speedup achieved by Multiverse-32B with varying degrees of parallelism across different batch sizes, while keeping a fixed 4K output length. The results in Figure~\ref{fig:speedup_vs_bsz} indicate that the generation process remains memory-bound as the batch size increases from 1 to 128. Therefore, the speedup of Multiverse scales linearly with the degree of parallelism across multiple configurations, showcasing its excellent scalability.

%% file: text/conclusion.tex
\section{Conclusion}
\label{Conclusion}
This work proposes Multiverse, a natively parallel generative model based on a MapReduce paradigm that internalizes three stages: (i) a \texttt{Map} stage for adaptive task decomposition, (ii) a \texttt{Process} stage for parallel subtask execution, and (iii) a \texttt{Reduce} stage for lossless result synthesis. To build a real-world Multiverse model, we co-design our data, algorithm, and system, enabling a seamless and rapid transfer from AR-LLMs. After fine-tuning on Multiverse-1K, our Multiverse-32B achieves performance comparable to AR-LLMs on real-world reasoning tasks, while achieving better scaling using the same context length due to parallel generation. Moreover, it leads to up to 2$\times$ wall-clock speedup per generated token across varying batch sizes, based on the degree of parallelism. \mbox{We hope that Multiverse can serve as an alternative to Autoregression for generative modeling.}

\section{Limitations}

While Multiverse provides a general framework for generative modeling, its application to diverse data and task types beyond LLM reasoning remains underexplored. Moreover, as Multiverse-32B was trained solely using Supervised Fine-Tuning (SFT), a key direction in future research is to integrate Reinforcement Learning (RL) into training to explore and \mbox{encourage more parallelism, which in turn would require a more robust Multiverse engine.}

\section{Broader Impacts}

Multiverse significantly boosts GPU utilization by enabling massive parallel generation. This modeling framework is particularly beneficial for small-batch and long-context inference scenarios, leading to substantial reductions in latency and corresponding energy consumption. Furthermore, Multiverse enables economies of scale for difficult but parallelizable tasks, decreasing the time per task unit while maintaining near-constant overall latency, even as task complexity increases. This remarkable scalability showcases its potential to address extremely complex tasks in practice that were \mbox{previously intractable, offering a promising path towards artificial superintelligence (ASI).}

\section*{Acknowledgement}

We thank Zhuoming Chen, Haizhong Zheng, Ranajoy Sadhukhan, Yang Zhou, Songlin Yang, Liliang Ren, Wentao Guo, Ruijie Zhu, Yu Zhang, and Yixin Dong for their constructive feedback on this work, along with the authors of s1~\citep{muennighoff2025s1}, SGLang~\citep{zheng2023efficiently}, and LightEval~\citep{lighteval}  for their useful codebase. We are also grateful to BitDeer AI Research for providing GPU resources and to Google for supplying Gemini API credits. This research was supported in part by a DGX B200 gift from NVIDIA, a Google Research Award, an Amazon Research Award, \mbox{Intel, Li Auto, Moffett AI, and the CMU CyLab Seed Fund.}

%% file: text/appendix.tex
\section{Prompt of Multiverse Curator}
\label{app:prompt}

In this section, we release our complete five-stage prompting protocol to create Multiverse-1K, powered by the Gemini 2.5 Pro model. This protocol is engineered to transform any sequential CoT data into Multiverse data.

This protocol starts with a multi-round conversation with the LLM (Stages 1-3) to convert an original reasoning chain into a parallel-structured summary. In Stage 4, both this summary and the original reasoning trajectory are fed to the LLM to repopulate each summarized step with its complete, original details. A content checker then immediately assesses these refilled steps. If the \textit{editor distance} (e.g., Levenshtein distance between the original trajectory ($s_{ori}$) and its rewritten version ($s_{gen}$), denoted as $d(s_{ori}, s_{gen})$) is too high, that step is re-generated. To normalize this, a \textit{relative editor distance} is calculated to \mbox{decide if a threshold $r$ is exceeded (set to $0.2$ in practice):}
$$
\text{Relative Editor Distance} = \frac{d(s_{ori}, s_{gen})}{\max(\text{length}(s_{ori}), \text{length}(s_{gen}))}
$$

Next, in Stage 5, we transform the output from Stage 4 into a MapReduce-structured reasoning trajectory by inserting the \texttt{Map} and \texttt{Reduce} phases that are generated by Gemini 2.5 Pro. To ensure the structural validity of the data, we perform a grammar check using a customized XML interpreter, which filters out invalid entries and extracts the outermost MapReduce blocks in the remaining valid ones. Finally, each path is rewritten separately to produce fully independent reasoning paths. The prompts used in the entire protocol are as follows:

\begin{tcolorbox}[title={STAGE 1: Generating a Summary Tree}, width=\textwidth, fontupper=\small]
{\normalsize\textbf{Main‐Step Extraction}}

Analyze the given reasoning chain (for a math or coding problem) and pull out every \textbf{major} step. Ignore substeps—only list the top‑level insights or actions.
\vspace{1em}

\textit{Output format} 
\begin{verbatim}
S1: [First major step]
S2: [Second major step]
S3: [Third major step]
…
SX: [Description of step X]
…
\end{verbatim}
\vspace{1em}

\textit{Guidelines}
\begin{itemize}[itemsep=0.0ex,topsep=0pt,leftmargin=*]
    \item 

 Label each top‑level step consecutively (`S1', `S2', `S3', …).  

\item 
 Please capture the entire thought process presented in the reasoning chain, and do not skip any step that includes but not is limited to:  
\begin{enumerate}
    \item 
     Initial problem understanding and analysis
  \item 
   All exploration paths (both successful and unsuccessful) 
  \item 
   Case studies, checks, or tests performed
  \item 
   Any “aha” or correction (re-evaluation or re-thinking) moments  
  \item 
   The final reasoning that yields the solution   
  \end{enumerate}
\item 
 Keep each item concise yet descriptive.  

\item 
Do \textbf{not} include any sub‑numbering (no `S2.1', etc.).
\item 
 Explicitly split multiple cases or scenarios into different steps. Each case should be allocated an independent step.

\end{itemize}

\vspace{2em}
{\normalsize\textbf{Substep Extraction}}

Given the output including all main step from a reasoning chain, break it down into all its internal substeps only if it can be meaningfully subdivided into smaller thought units.
\vspace{1em}

\textit{Output format}
\begin{verbatim}
S1: [Description of step 1]
S2: [Description of step 2]
    S2.1 [Description of step 2.1]
    S2.2 [Description of step 2.2]
    …
    S2.10 [Description of step 2.10]
S3: [Description of step 3]
S4: [Description of step 4]
…
S10: [Description of step 10]
…
\end{verbatim}
\vspace{1em}

\textit{Guidelines}
\begin{itemize}[itemsep=0.0ex,topsep=0pt,leftmargin=*]
  \item Use the same parent index (`x') as the main step (e.g.\ if breaking down `S2', label `S2.1', `S2.2', …).
  \item Capture the entire thought process presented in the reasoning chain, and do not skip any substep that includes but is not limited to:
    \begin{enumerate}[itemsep=0.0ex,topsep=0pt,leftmargin=*]
      \item Initial problem understanding and analysis
      \item All exploration paths (both successful and unsuccessful)
      \item Case studies, checks, or tests performed
      \item Any “aha” or correction (re-evaluation or re-thinking) moments
      \item The final reasoning that yields the solution
    \end{enumerate}
  \item Do \textbf{not} introduce deeper nesting larger than 2 (e.g.\ `S2.1.1' is not allowed).
  \item Explicitly split multiple cases or scenarios into different substeps. Each case should be allocated an independent substep.
\end{itemize}
\end{tcolorbox}

\begin{tcolorbox}[title={STAGE 2: Identifying Parallel Groups}, breakable, width=\textwidth, fontupper=\small]
{\normalsize\textbf{Parallelizing Main Steps}}

Using only the  \textbf{main steps} (\verb|S1|, \verb|S2|, …) you extracted in Stage 1, identify all steps or contiguous step groups that can be executed in parallel without violating logical dependencies, and rewrite the plan as a structured parallel execution outline.
\vspace{1em}

\textit{1. Identify Parallel Groups}
\begin{itemize}[itemsep=0.0ex,topsep=0pt,leftmargin=*]
  \item Find sets of adjacent main steps with no dependencies among them.
  \item Label groups \verb|P1|, \verb|P2|, … and list their step ranges (e.g.\ \verb|[S1+S2, S3]|, \verb|[S4]|).
\end{itemize}

\textit{2. Rewrite into a Parallel Execution Plan}
\begin{itemize}[itemsep=0.0ex,topsep=0pt,leftmargin=*]
  \item Preserve each step’s original wording as much as possible.
\end{itemize}

\vspace{1em}
\textit{Output Format:}
\begin{verbatim}
Parallel groups:
P1: [S1+S2, S3]
P2: [S4]
...

Parallel execution plan:
P1[parallel reason: ...]:
  S1+S2: [text of S1 + text of S2]
  S3: [text of S3]
P2[parallel reason: ...]:
  S4: [text of S4]
...
\end{verbatim}
\vspace{1em}

\textit{Guidelines}
\begin{itemize}[itemsep=0.0ex,topsep=0pt,leftmargin=*]
  \item \textbf{Coverage}: Include \textbf{every} step exactly once, either alone or inside a parallel group.
  \item \textbf{Contiguous Blocks}: Combine only adjacent steps into blocks; do \textbf{not} combine non-adjacent steps.
  \item \textbf{Strict Parallelism Only}: Build a dependency graph: draw an edge from step A to B if B uses A’s output. A group \verb|P_i| may include steps (or blocks) only if there are no edges between them. Treat conditional branches as independent tasks.
 \item \textbf{Contiguous Grouping Only}: Each parallel group must cover a continuous sequence of steps. Do not parallelize non-adjacent steps.
  \item \textbf{Conciseness}: Keep each bullet short and stick closely to the original text.
\end{itemize}
\vspace{2em}

{\normalsize\textbf{Parallelizing Substeps}}

Using only the \textbf{substeps} (\texttt{S2.1}, \texttt{S2.2}, ...) you extracted in Stage 1, identify all substeps or contiguous substep groups can be executed in parallel without violating logical dependencies, and rewrite the plan as a structured parallel execution outline.
\vspace{1em}

\textit{1. Identify Parallel Groups}
\begin{itemize}[itemsep=0.0ex,topsep=0pt,leftmargin=*]
  \item Find sets of adjacent main steps with no dependencies among them.
  \item Label groups \verb|P1|, \verb|P2|, … and list their step ranges (e.g.\ \verb|[S2.1+S2.2, S2.3]|, \verb|[S3.1]|).
\end{itemize}

\textit{2. Rewrite into a Parallel Execution Plan}
\begin{itemize}[itemsep=0.0ex,topsep=0pt,leftmargin=*]
  \item Preserve each step’s original wording as much as possible.
\end{itemize}

\vspace{1em}
\textit{Output Format:}
\begin{verbatim}
Parallel groups:
P1: [S2.1+S2.2, S2.3]
P2: [S2.4]
P2: [S3.1]
...

Parallel execution plan:
P1[parallel reason: ...]:
  S2.1+S2.2: [text of S2.1 + text of S2.2]
  S2.3: [text of S2.3]
P2[parallel reason: ...]:
  S3.1: [text of S3.1]
...
\end{verbatim}
\vspace{1em}

\textit{Guidelines}
\begin{itemize}[itemsep=0.0ex,topsep=0pt,leftmargin=*]
  \item \textbf{Coverage}: Include \textbf{every} substep exactly once, either alone or inside a parallel group.
  \item \textbf{Contiguous Blocks}: Combine only adjacent substeps into blocks; do \textbf{not} combine non-adjacent substeps.
  \item \textbf{Strict Parallelism Only}: Build an explicit dependency graph in your analysis: draw an edge from substep A to substep B if B uses A’s output or insight. A group \verb|Pi| may include steps (or contiguous blocks) only if there are no edges between any two steps. In conditional logic, treat the \textbf{if} branch and \textbf{else} branch as independent tasks and parallelize them even though their outputs cannot both occur at runtime.
 \item \textbf{Contiguous Grouping Only}: Each parallel group must cover a continuous sequence of steps or blocks. In other words, you may only parallelize adjacent substeps. The occurrence of substeps in parallel groups must follow their original order. For example, \verb|P1: [S2.2, S3.1]| is not allowed.
  \item \textbf{Conciseness}: Keep each bullet short and stick closely to the original text.
\end{itemize}
\end{tcolorbox}
\vspace{2em}

\begin{tcolorbox}[title={STAGE 3: Reformating into Parallel Structures}, width=\textwidth, fontupper=\small]

{\normalsize\textbf{Get Structured Summary}}

Please summarize the conversation above by extracting the reasoning steps and substeps in Stage 1 as a tree structure with explicit parallelism annotations following Stage 2.
\vspace{1em}

\textit{Output Format}
\begin{verbatim}
O1: [Brief summary of top-level step S1] 
<parallel>[parallel reason: ...]
O1.1: [Summary of substep S1.1]
O1.2: [Summary of substep S1.2]
… 
</parallel> 
<parallel>[parallel reason: ...]
O2: [Brief description of top-level step S2]
<parallel>[parallel reason: ...]
O2.1: [Summary of substep S2.1 + Summary of substep S2.2]
O2.2: [Summary of substep S2.3]
… 
</parallel> 
O3: [Brief description of top-level step S3] 
</parallel>
O4: [Brief description of top-level step S4]
…
\end{verbatim}
\vspace{1em}

\textit{Guidelines}
\begin{itemize}[itemsep=0.0ex,topsep=0pt,leftmargin=*]
\item \textbf{Max depth of nested \texttt{<parallel>} is 2.} Do not nest \texttt{<parallel>} tags more deeply than two levels.
\item \textbf{Max depth of nested numbering is 2.} Only use \texttt{Ox} and \texttt{Ox.y}; do not introduce deeper numbering like \texttt{Ox.y.z}.
\item \textbf{Sequential subpaths stay unexpanded.} If a node’s children are purely sequential, list them normally without any \texttt{<parallel>} wrapper.
\item \textbf{Tag parallel blocks.} Wrap only genuinely parallelizable sibling steps in a \texttt{<parallel>}…\texttt{</parallel>} block, and include a parallel‐reason annotation.
\item \textbf{Concise summaries.} Each step and substep should be described briefly and clearly.
\item \textbf{Avoid over-splitting.} If most children are sequential and only a pair can run in parallel, either leave the group un-split or tag only the truly parallel pair.
\item \textbf{Group parallelizable sets.} You may combine several independent paths into one \texttt{<parallel>} block when they share no dependencies.
\end{itemize}

\end{tcolorbox}

\begin{tcolorbox}[title={STAGE 4: Refilling Original Details}, width=\textwidth, fontupper=\small]
{\normalsize\textbf{Refill the Full, Detailed Reasoning Trajectories into the Structured Summary}}

You will receive an outline that \emph{may be incomplete but includes \texttt{<parallel>} tags} indicating parallel structures.  It contain summaries for several steps and substeps.  You will also receive the corresponding original text, where sentences implicitly or explicitly map to hierarchical prefixes (e.g., \verb|O1|, \verb|O1.1|, \verb|O2|) in sequence. Your task is to process the original reasoning chain sequentially to update the outline: replace existing summaries or insert new steps as needed, while preserving the original \verb|<parallel>| tag structure.
\vspace{1em}

\textit{Guidelines:}
\begin{itemize}[itemsep=0.0ex,topsep=0pt,leftmargin=*]
  \item \textbf{Initialize Structure}
    Start with the structure provided by the input outline, including its text/summaries and all \verb|<parallel>| tags in their original locations.
  \item \textbf{Read Sentences Sequentially:}
    Process each sentence of the original text one by one, in the exact order they appear.
  \item \textbf{Process Each Sentence:}
    \begin{enumerate}[itemsep=0.0ex,topsep=0pt,leftmargin=*]
      \item Determine the hierarchical prefix associated with this sentence (e.g., \verb|O1|, \verb|O1.1|, \verb|O2|).
      \item Check if a step or substep with this prefix already exists in the outline.
      \item \emph{If it exists:}  
        Replace its current summary with the full original sentence.
      \item \emph{If it does not exist:}  
        Insert a new step/substep at the correct hierarchical position (e.g., \verb|S1.1| under \verb|S1|, \verb|S2| after \verb|S1|), using the full original sentence as its content and matching the outline’s indentation.
    \end{enumerate}
  \item \textbf{Preserve \texttt{<parallel>} Tags:}
    Keep every existing \verb|<parallel>| and \verb|</parallel>| tag exactly where it was in the input outline.  Do not add, remove, or relocate any tags.
  \item \textbf{Ensure Correct Output Formatting:}
    \begin{itemize}[itemsep=0.0ex,topsep=0pt,leftmargin=*]
      \item Maintain proper hierarchical indentation for all steps and substeps.
      \item Each entry must be on its own line, beginning with its prefix (e.g., \verb|O1:|, \verb|O1.1:|), followed by the full original sentence.
    \end{itemize}
  \item \textbf{Maintain Completeness:}  
    Verify that every sentence from the original reasoning chain has been processed and appears in the updated outline.  Do not omit or merge any sentences.
\end{itemize}
\end{tcolorbox}
\vspace{2em}

\begin{tcolorbox}[title={STAGE 5: Adding MapReduce Structures \& Rewriting All Paths}, breakable, width=\textwidth, fontupper=\small]
{\normalsize\textbf{Filling Detailed Goal and Conclusion Based on the New Reasoning Trajectory}}

Based on the generated reasoning chain, your task is to transform it according to the following rules:
\vspace{1em}

\textit{Output Format}
\begin{verbatim}

[Full reasoning copied from the reasoning chain for the first top-level path]
[Full reasoning copied from the reasoning chain for the second top-level path]
...
Let's think in parallel.
<Parallel>
<Goal>
Path: [brief, self-contained description of case A]
Path: [brief, self-contained description of case B]
...
</Goal>
<Path>
[Introductory reasoning for case A]
Let's think in parallel.
<Parallel>
<Goal>
Path: [brief, self-contained description of case A.1]
Path: [brief, self-contained description of case A.2]
</Goal>
<Path>
[Full detailed reasoning for case A.1, rewritten clearly and independently]
</Path>
<Path>
[Full detailed reasoning for case A.2, rewritten clearly and independently]
</Path>
<Conclusion>
[Your concise summary of outcomes from A.1 and A.2]
</Conclusion>
</Parallel>
</Path>
<Path>
[Full detailed reasoning for case B, rewritten clearly and independently]
</Path>
...
<Conclusion>
[Your concise summary of outcomes from A and B]
</Conclusion>
</Parallel>
[Full detailed reasoning for any remaining paths]
\end{verbatim}
\vspace{1em}

\textit{Guidelines}
\begin{itemize}[itemsep=0.0ex,topsep=0pt,leftmargin=*]
    \item                                               Remove all numbering labels (e.g., \texttt{O1}, \texttt{O2.1}) and eliminate any indentation.
    \item For each \verb|<Parallel>…</Parallel>| block: 
      \begin{itemize}
        \item Group every step, substep, and subsubstep belonging to the same parallel branch into a single \verb|<Path>…</Path>| section.
        \item Discard the \verb|[parallel reason: …]| annotations.
      \end{itemize}
    \item Within each \verb|<Parallel>| block:
      \begin{itemize}
        \item Insert \verb|<Goal>| before the first \verb|<Path>|, listing each branch as \texttt{Path:} … .
        \item Insert \verb|<Conclusion>| after the last \verb|<Path>|, summarizing each branch’s outcome independently.
      \end{itemize}
    \item When multiple \verb|<Path>| entries stem from the same original sentence or have interdependencies:
      \begin{itemize}
        \item Rewrite each path separately and completely, ensuring no cross-references.
        \item Provide enough context in each \verb|<Path>| so it stands alone.
        \item Fully encapsulate the logical reasoning for each path.
      \end{itemize}
    \item Avoid repetition: do not echo the brief descriptions from \verb|<Goal>| inside the corresponding \verb|<Path>|, and minimize redundant information across paths.
\end{itemize}
\vspace{2em}
{\normalsize\textbf{Rewriting Paths in the Structured Reasoning Trajectory}}

You are given a full structured reasoning trajectory inside a \texttt{<Parallel>} block, consisting of:

\begin{itemize}[itemsep=0.0ex,topsep=0pt,leftmargin=*]
\item one \texttt{<Goal>} block with multiple \texttt{<Outline>} elements
\item multiple \texttt{<Path>} blocks
\item one \texttt{<Conclusion>} block.
\end{itemize}

Some \texttt{<Path>} blocks may contain an entire nested \texttt{<Parallel>} structure (from \texttt{<Parallel>} to \texttt{</Parallel>}). These nested blocks should be rewritten using the same rules recursively.
\vspace{0.5em}

\textbf{For \texttt{<Goal>}}:
\begin{itemize}[itemsep=0.0ex,topsep=0pt,leftmargin=*]
\item Rewrite each \texttt{<Outline>} into a \textbf{concise statement of what is being calculated or determined}.
\item Remove any content describing \textbf{how} the problem is solved or intermediate reasoning steps.
\end{itemize}
\vspace{0.5em}

\textbf{For each \texttt{<Path>}}:
\begin{itemize}[itemsep=0.0ex,topsep=0pt,leftmargin=*]
\item Keep the original numbering prefix (e.g., `1:', `2:').
\item Rewrite the content as a \textbf{complete, fluent, and logically self-contained paragraph}.
\item Do \textbf{not} use transitional phrases like ``First," ``Then," ``Next," ``On the other hand," etc.
\item If the \texttt{<Path>} contains \textbf{five or fewer sentences}, rewrite them together as a single coherent paragraph, ensuring logical flow and fluency without using transitional phrases.
\item If the \texttt{<Path>} contains \textbf{more than five sentences}: Rewrite the first five sentences together as a single unit, forming a fluent paragraph. For the remaining sentences, rewrite each one individually, based on its meaning, as clear and fluent standalone statements.
\item If the \texttt{<Path>} contains a \textbf{nested \texttt{<Parallel>} block}, apply all these rules recursively to the nested block.
\end{itemize}

Each \texttt{<Path>} must make sense independently, even if it contains a nested reasoning chain.
\vspace{1em}

\textbf{For \texttt{<Conclusion>}}:
\begin{itemize}[itemsep=0.0ex,topsep=0pt,leftmargin=*]
\item Rewrite the conclusion as the \textbf{most concise and synthesized summary} of the main outcomes from all \texttt{<Path>} blocks.
\item You may combine or compare results from different paths, but keep it succinct and direct.
\end{itemize}

\vspace{0.5em}

\textbf{Nested \texttt{<Parallel>}:}

\begin{itemize}[itemsep=0.0ex,topsep=0pt,leftmargin=*]
\item A nested \texttt{<Parallel>} may appear only \textbf{as a full block inside a \texttt{<Path>}}.
\item If a \texttt{<Path>} contains a nested \texttt{<Parallel>...</Parallel>} block, process that inner block exactly as you would the top-level one:
\begin{itemize}[itemsep=0.0ex,topsep=0pt,leftmargin=*]
  \item Rewrite the inner \texttt{<Goal>}, \texttt{<Path>}, and \texttt{<Conclusion>} elements accordingly.
  \item Maintain the XML structure — do not reindent or alter the tag hierarchy.
\end{itemize}
\end{itemize}
\vspace{1em}

\textit{Output Format}
\begin{verbatim}
<Parallel>
<Goal>
<Outline>
1: [concise description of the goal of Path 1]
</Outline>
<Outline>
2: [concise description of the goal of Path 2]
</Outline>
</Goal>
<Path>
1: [self-contained paragraph for Path 1, rewritten sentence by sentence]
<Parallel>
<Goal>
<Outline>
1.1: [concise goal for Path 1.1]
</Outline>
<Outline>
1.2: [concise goal for Path 1.2]
</Outline>
</Goal>
<Path>
1.1: [rewritten paragraph for Path 1.1]
</Path>
<Path>
1.2: [rewritten paragraph for Path 1.2]
</Path>
<Conclusion>
[summarized results from Paths 1.1 and 1.2]
</Conclusion>
</Parallel>
</Path>
<Path>
2: [self-contained paragraph for Path 2, rewritten sentence by sentence]
</Path>
<Conclusion>
[summary of outcomes from Path 1 and Path 2]
</Conclusion>
</Parallel>
\end{verbatim}
\vspace{1em}

\textit{Guidelines}
\begin{itemize}[itemsep=0.0ex,topsep=0pt,leftmargin=*]
\item You must preserve all XML tags exactly as they appear — do not remove, rename, or reorder tags.
\item Only modify the inner text of \texttt{<Outline>}, \texttt{<Path>}, and \texttt{<Conclusion>} blocks.
\item Wrap the final output in a single Markdown code block using triple backticks.
\end{itemize}

\end{tcolorbox}

\section{Examples in Training Data and Model Generation}
We present several examples from our Multiverse-1K dataset, along with the corresponding outputs generated by Multiverse-32B, which was trained on this dataset. Both typical collective and selective branches are included.

\subsection{Training Examples from Multiverse-1K}
\begin{tcolorbox}[title={Multiverse Data A: Collective Case}, breakable, width=\textwidth, fontupper=\small]
\texttt{<Parallel>}

\texttt{<Goal>}

\texttt{<Outline>}

1: Determine the sum of distances from $P_{1}$ to $P_{2}$, $P_{3}$, and $P_{4}$.

\texttt{</Outline>}

\texttt{<Outline>}

2: Determine the sum of distances from $P_{2}$ to $P_{1}$, $P_{3}$, and $P_{4}$.

\texttt{</Outline>}

\texttt{<Outline>}

3: Determine the sum of distances from $P_{3}$ to $P_{1}$, $P_{2}$, and $P_{4}$.

\texttt{</Outline>}

\texttt{<Outline>}

4: Determine the sum of distances from $P_{4}$ to $P_{1}$, $P_{2}$, and $P_{3}$.

\texttt{</Outline>}

\texttt{</Goal>}

\texttt{<Path>}

1: For point $P_{1}(0,0)$,  
the distance to $P_{2}(10,20)$ is  
$$\sqrt{(10-0)^{2} + (20-0)^{2}} = \sqrt{100 + 400} = \sqrt{500} \approx 22.36.$$  
The distance to $P_{3}(5,15)$ is  
$$\sqrt{(5-0)^{2} + (15-0)^{2}} = \sqrt{25 + 225} = \sqrt{250} \approx 15.81.$$  
The distance to $P_{4}(12,-6)$ is  
$$\sqrt{(12-0)^{2} + (-6-0)^{2}} = \sqrt{144 + 36} = \sqrt{180} \approx 13.42.$$  
The sum is $22.36 + 15.81 + 13.42 \approx 51.59$.

\texttt{</Path>}

\texttt{<Path>}

2: For point $P_{2}(10,20)$,  
the distance to $P_{1}(0,0)$ is  
$$\sqrt{(10-0)^{2} + (20-0)^{2}} = \sqrt{500} \approx 22.36.$$  
The distance to $P_{3}(5,15)$ is  
$$\sqrt{(10-5)^{2} + (20-15)^{2}} = \sqrt{25 + 25} = \sqrt{50} \approx 7.07.$$  
The distance to $P_{4}(12,-6)$ is  
$$\sqrt{(10-12)^{2} + (20-(-6))^{2}} = \sqrt{4 + 676} = \sqrt{680} \approx 26.08.$$  
The sum is $22.36 + 7.07 + 26.08 \approx 55.51$.

\texttt{</Path>}

\texttt{<Path>}

3: For point $P_{3}(5,15)$,  
the distance to $P_{1}(0,0)$ is  
$$\sqrt{(5-0)^{2} + (15-0)^{2}} = \sqrt{250} \approx 15.81.$$  
The distance to $P_{2}(10,20)$ is  
$$\sqrt{(5-10)^{2} + (15-20)^{2}} = \sqrt{50} \approx 7.07.$$  
The distance to $P_{4}(12,-6)$ is  
$$\sqrt{(5-12)^{2} + (15-(-6))^{2}} = \sqrt{49 + 441} = \sqrt{490} \approx 22.14.$$  
The sum is $15.81 + 7.07 + 22.14 \approx 45.02$.

\texttt{</Path>}

\texttt{<Path>}

4: For point $P_{4}(12,-6)$,  
the distance to $P_{1}(0,0)$ is  
$$\sqrt{(12-0)^{2} + (-6-0)^{2}} = \sqrt{180} \approx 13.42.$$  
The distance to $P_{2}(10,20)$ is  
$$\sqrt{(12-10)^{2} + (-6-20)^{2}} = \sqrt{4 + 676} = \sqrt{680} \approx 26.08.$$  
The distance to $P_{3}(5,15)$ is  
$$\sqrt{(12-5)^{2} + (-6-15)^{2}} = \sqrt{49 + 441} = \sqrt{490} \approx 22.14.$$  
The sum is $13.42 + 26.08 + 22.14 \approx 61.64$.

\texttt{</Path>}

\texttt{<Conclusion>}

The sums of distances from $P_{1}, P_{2}, P_{3}, P_{4}$ to the other three points are approximately $51.59$, $55.51$, $45.02$, and $61.64$, respectively.

\texttt{</Conclusion>}

\texttt{</Parallel>}

\end{tcolorbox}
\vspace{2em}

\begin{tcolorbox}[title={Multiverse Data B: Selective Case}, breakable, width=\textwidth, fontupper=\small]
\texttt{<Parallel>}

\texttt{<Goal>}

\texttt{<Outline>}

1: Determine if $H$ can be constructed as a countable intersection of open dense sets using Baire Category Theorem (BCT), leveraging the completeness of $S^2$.

\texttt{</Outline>}

\texttt{<Outline>}

2: Determine if $H$ can be constructed step-by-step using $\Omega$’s approximation property to ensure density and $K$-avoidance for chords, while addressing potential difficulties.

\texttt{</Outline>}

\texttt{</Goal>}

\texttt{<Path>}

1: The sphere $S^2$ is a complete metric space, suggesting the possibility of constructing $H$ as a countable intersection of open dense sets; however, the specific method for such a construction is not immediately clear.

\texttt{</Path>}

\texttt{<Path>}

2: A step-by-step construction of $H$ could leverage the given condition to ensure density and that all chords avoid $K$; $H$ must be dense in $S^2$, meaning it has points in every open subset of $S^2$, and any two points in $H$ must have their connecting chord disjoint from $K$. The family $\Omega$ allows approximation of any two points on $S^2$ with a chord $X'Y'$ from $\Omega$. This implies $\Omega$ is dense in the space of all chords, with the additional property that its chords avoid $K$. The closed nature of $K$ means its complement within the unit ball is open, so a chord disjoint from $K$ lies entirely within this open complement. Consequently, $\Omega$ comprises chords in the complement of $K$. The challenge is to construct a dense set $H$ such that all chords connecting points within $H$ avoid $K$, given that for any $X,Y\in S^2$, nearby points $X',Y'$ exist whose chord $X'Y'$ is in $\Omega$ and thus avoids $K$. One approach might be to construct $H$ as a countable dense set where each pair’s connecting chord belongs to $\Omega$. Achieving density alone is insufficient; all chords between any two points in $H$ must be in $\Omega$. Directly constructing such an $H$ presents significant difficulties.

\texttt{</Path>}

\texttt{<Conclusion>}

The Baire Category Theorem approach for constructing $H$ via intersections (Path 1) lacks a clear method. A step-by-step construction (Path 2) highlights the requirements of density and $K$-avoidance for all chords within $H$, but faces challenges in maintaining the chord property universally during the construction process.

\texttt{</Conclusion>}

\texttt{</Parallel>}
\end{tcolorbox}

\subsection{Generated Examples from Multiverse-32B}

\begin{tcolorbox}[title={Multiverse Generation A: Collective Case}, breakable, width=\textwidth, fontupper=\small]
    \texttt{<Parallel>}

\texttt{<Goal>}

\texttt{<Outline>}

1: Compute Case 1 count.

\texttt{</Outline>}

\texttt{<Outline>}

2: Compute Case 2 count.

\texttt{</Outline>}

\texttt{</Goal>}

\texttt{<Path>}

1: In Case 1 the path pattern is H–V–H–V–H, giving three horizontal and two vertical segments.  
We distribute $8$ units into three positive horizontal segments:  
\[
\binom{8-1}{3-1} \;=\;\binom{7}{2}\;=\;21.
\]  
We distribute $8$ units into two positive vertical segments:  
\[
\binom{8-1}{2-1} \;=\;\binom{7}{1}\;=\;7.
\]  
Total for Case 1:  
\[
21 \times 7 \;=\;147.
\]

\texttt{</Path>}

\texttt{<Path>}

2: In Case 2 the pattern is V–H–V–H–V, giving three vertical and two horizontal segments.  
Distribute $8$ units into three positive vertical segments:  
\[
\binom{7}{2} = 21.
\]  
Distribute $8$ units into two positive horizontal segments:  
\[
\binom{7}{1} = 7.
\]  
Total for Case 2:  
\[
21 \times 7 = 147.
\]

\texttt{</Path>}

\texttt{<Conclusion>}

Case 1 yields $147$ paths; Case 2 yields $147$ paths; overall $147 + 147 = 294$.

\texttt{</Conclusion>}

\texttt{</Parallel>}

\end{tcolorbox}

\vspace{2em}

\begin{tcolorbox}[title={Multiverse Generation B: Selective Case}, , breakable, width=\textwidth, fontupper=\small]
\texttt{<Parallel>}

\texttt{<Goal>}

\texttt{<Outline>}

1: Count horizontal vs. vertical segments.

\texttt{</Outline>}

\texttt{<Outline>}

2: List possible H/V patterns.

\texttt{</Outline>}

\texttt{</Goal>}

\texttt{<Path>}

1: A path from $(0,0)$ to $(8,8)$ uses exactly eight right moves ($R$) and eight up moves ($U$). With five segments alternating direction, starting with $R$ forces the pattern  
$$R\;-\;U\;-\;R\;-\;U\;-\;R,$$  
giving three horizontal segments and two vertical segments. Conversely, starting with $U$ yields  
$$U\;-\;R\;-\;U\;-\;R\;-\;U,$$  
for three vertical and two horizontal segments. In either case there are four direction changes (one at each segment boundary), and the segment counts differ by one because there are an odd number of segments.

\texttt{</Path>}

\texttt{<Path>}

2: The total move count is $16$ ($8R + 8U$). If $H=3$ and $V=2$, then the three horizontal segment lengths $h_{1}+h_{2}+h_{3}=8$ with each $h_{i}\ge1$, which has  
$$\binom{8-1}{3-1}=\binom{7}{2}=21$$  
compositions. The two vertical segments $v_{1}+v_{2}=8$ with $v_{j}\ge1$ has  
$$\binom{8-1}{2-1}=\binom{7}{1}=7$$  
compositions, for $21\times7=147$ ways when starting horizontal. Similarly, if $V=3$ and $H=2$, there are $21$ ways to split $8$ into three vertical segments and $7$ ways for two horizontal segments, again $147$ ways. In total there are $147+147=294$ segment‐length configurations consistent with five alternating segments summing to $(8,8)$.

\texttt{</Path>}

\texttt{<Conclusion>}

A five‐segment path must alternate directions, yielding either $(H,V)=(3,2)$ or $(2,3)$ segments. There are $147$ ways to allocate segment lengths in each case, for $294$ total valid patterns.

\texttt{</Conclusion>}

\texttt{</Parallel>}

\end{tcolorbox}

%% file: main.bbl
\begin{thebibliography}{49}
\providecommand{\natexlab}[1]{#1}
\providecommand{\url}[1]{\texttt{#1}}
\expandafter\ifx\csname urlstyle\endcsname\relax
  \providecommand{\doi}[1]{doi: #1}\else
  \providecommand{\doi}{doi: \begingroup \urlstyle{rm}\Url}\fi

\bibitem[Aho et~al.(1974)Aho, Hopcroft, and Ullman]{aho1974design}
Alfred~V. Aho, John~E. Hopcroft, and Jeffrey~D. Ullman.
\newblock \emph{The Design and Analysis of Computer Algorithms}.
\newblock Addison-Wesley, 1974.
\newblock ISBN 9780201000290.

\bibitem[Arriola et~al.(2025)Arriola, Gokaslan, Chiu, Yang, Qi, Han, Sahoo, and Kuleshov]{arriola2025block}
Marianne Arriola, Aaron Gokaslan, Justin~T Chiu, Zhihan Yang, Zhixuan Qi, Jiaqi Han, Subham~Sekhar Sahoo, and Volodymyr Kuleshov.
\newblock Block diffusion: Interpolating between autoregressive and diffusion language models.
\newblock \emph{arXiv preprint arXiv:2503.09573}, 2025.

\bibitem[Austin et~al.(2021)Austin, Johnson, Ho, Tarlow, and Van Den~Berg]{austin2021structured}
Jacob Austin, Daniel~D Johnson, Jonathan Ho, Daniel Tarlow, and Rianne Van Den~Berg.
\newblock Structured denoising diffusion models in discrete state-spaces.
\newblock \emph{Advances in neural information processing systems}, 34:\penalty0 17981--17993, 2021.

\bibitem[Barrault et~al.(2024)Barrault, Duquenne, Elbayad, Kozhevnikov, Alastruey, Andrews, Coria, Couairon, Costa-juss{\`a}, Dale, et~al.]{barrault2024large}
Lo{\"\i}c Barrault, Paul-Ambroise Duquenne, Maha Elbayad, Artyom Kozhevnikov, Belen Alastruey, Pierre Andrews, Mariano Coria, Guillaume Couairon, Marta~R Costa-juss{\`a}, David Dale, et~al.
\newblock Large concept models: Language modeling in a sentence representation space.
\newblock \emph{arXiv preprint arXiv:2412.08821}, 2024.

\bibitem[Brown et~al.(2024)Brown, Juravsky, Ehrlich, Clark, Le, R{\'e}, and Mirhoseini]{brown2024large}
Bradley Brown, Jordan Juravsky, Ryan Ehrlich, Ronald Clark, Quoc~V Le, Christopher R{\'e}, and Azalia Mirhoseini.
\newblock Large language monkeys: Scaling inference compute with repeated sampling.
\newblock \emph{arXiv preprint arXiv:2407.21787}, 2024.

\bibitem[Cobbe et~al.(2021)Cobbe, Kosaraju, Bavarian, Chen, Jun, Kaiser, Plappert, Tworek, Hilton, Nakano, et~al.]{cobbe2021training}
Karl Cobbe, Vineet Kosaraju, Mohammad Bavarian, Mark Chen, Heewoo Jun, Lukasz Kaiser, Matthias Plappert, Jerry Tworek, Jacob Hilton, Reiichiro Nakano, et~al.
\newblock Training verifiers to solve math word problems.
\newblock \emph{arXiv preprint arXiv:2110.14168}, 2021.

\bibitem[Dean and Ghemawat(2008)]{dean2008mapreduce}
Jeffrey Dean and Sanjay Ghemawat.
\newblock Mapreduce: simplified data processing on large clusters.
\newblock \emph{Communications of the ACM}, 51\penalty0 (1):\penalty0 107--113, 2008.

\bibitem[Dean et~al.(2004)Dean, Ghemawat, et~al.]{dean2004mapreduce}
Jeffrey Dean, Sanjay Ghemawat, et~al.
\newblock Mapreduce: simplified data processing on large clusters.
\newblock In \emph{osdi}, volume~4, page~5. USA, 2004.

\bibitem[DeepSeek-AI et~al.(2025)DeepSeek-AI, Guo, Yang, Zhang, Song, Zhang, Xu, Zhu, Ma, Wang, Bi, Zhang, Yu, Wu, Wu, Gou, Shao, Li, Gao, Liu, Xue, Wang, Wu, Feng, Lu, Zhao, Deng, Zhang, Ruan, Dai, Chen, Ji, Li, Lin, Dai, Luo, Hao, Chen, Li, Zhang, Bao, Xu, Wang, Ding, Xin, Gao, Qu, Li, Guo, Li, Wang, Chen, Yuan, Qiu, Li, Cai, Ni, Liang, Chen, Dong, Hu, Gao, Guan, Huang, Yu, Wang, Zhang, Zhao, Wang, Zhang, Xu, Xia, Zhang, Zhang, Tang, Li, Wang, Li, Tian, Huang, Zhang, Wang, Chen, Du, Ge, Zhang, Pan, Wang, Chen, Jin, Chen, Lu, Zhou, Chen, Ye, Wang, Yu, Zhou, Pan, Li, Zhou, Wu, Ye, Yun, Pei, Sun, Wang, Zeng, Zhao, Liu, Liang, Gao, Yu, Zhang, Xiao, An, Liu, Wang, Chen, Nie, Cheng, Liu, Xie, Liu, Yang, Li, Su, Lin, Li, Jin, Shen, Chen, Sun, Wang, Song, Zhou, Wang, Shan, Li, Wang, Wei, Zhang, Xu, Li, Zhao, Sun, Wang, Yu, Zhang, Shi, Xiong, He, Piao, Wang, Tan, Ma, Liu, Guo, Ou, Wang, Gong, Zou, He, Xiong, Luo, You, Liu, Zhou, Zhu, Xu, Huang, Li, Zheng, Zhu, Ma, Tang, Zha, Yan, Ren, Ren, Sha, Fu, Xu, Xie, Zhang,
  Hao, Ma, Yan, Wu, Gu, Zhu, Liu, Li, Xie, Song, Pan, Huang, Xu, Zhang, and Zhang]{deepseekai2025deepseekr1incentivizingreasoningcapability}
DeepSeek-AI, Daya Guo, Dejian Yang, Haowei Zhang, Junxiao Song, Ruoyu Zhang, Runxin Xu, Qihao Zhu, Shirong Ma, Peiyi Wang, Xiao Bi, Xiaokang Zhang, Xingkai Yu, Yu~Wu, Z.~F. Wu, Zhibin Gou, Zhihong Shao, Zhuoshu Li, Ziyi Gao, Aixin Liu, Bing Xue, Bingxuan Wang, Bochao Wu, Bei Feng, Chengda Lu, Chenggang Zhao, Chengqi Deng, Chenyu Zhang, Chong Ruan, Damai Dai, Deli Chen, Dongjie Ji, Erhang Li, Fangyun Lin, Fucong Dai, Fuli Luo, Guangbo Hao, Guanting Chen, Guowei Li, H.~Zhang, Han Bao, Hanwei Xu, Haocheng Wang, Honghui Ding, Huajian Xin, Huazuo Gao, Hui Qu, Hui Li, Jianzhong Guo, Jiashi Li, Jiawei Wang, Jingchang Chen, Jingyang Yuan, Junjie Qiu, Junlong Li, J.~L. Cai, Jiaqi Ni, Jian Liang, Jin Chen, Kai Dong, Kai Hu, Kaige Gao, Kang Guan, Kexin Huang, Kuai Yu, Lean Wang, Lecong Zhang, Liang Zhao, Litong Wang, Liyue Zhang, Lei Xu, Leyi Xia, Mingchuan Zhang, Minghua Zhang, Minghui Tang, Meng Li, Miaojun Wang, Mingming Li, Ning Tian, Panpan Huang, Peng Zhang, Qiancheng Wang, Qinyu Chen, Qiushi Du, Ruiqi Ge, Ruisong
  Zhang, Ruizhe Pan, Runji Wang, R.~J. Chen, R.~L. Jin, Ruyi Chen, Shanghao Lu, Shangyan Zhou, Shanhuang Chen, Shengfeng Ye, Shiyu Wang, Shuiping Yu, Shunfeng Zhou, Shuting Pan, S.~S. Li, Shuang Zhou, Shaoqing Wu, Shengfeng Ye, Tao Yun, Tian Pei, Tianyu Sun, T.~Wang, Wangding Zeng, Wanjia Zhao, Wen Liu, Wenfeng Liang, Wenjun Gao, Wenqin Yu, Wentao Zhang, W.~L. Xiao, Wei An, Xiaodong Liu, Xiaohan Wang, Xiaokang Chen, Xiaotao Nie, Xin Cheng, Xin Liu, Xin Xie, Xingchao Liu, Xinyu Yang, Xinyuan Li, Xuecheng Su, Xuheng Lin, X.~Q. Li, Xiangyue Jin, Xiaojin Shen, Xiaosha Chen, Xiaowen Sun, Xiaoxiang Wang, Xinnan Song, Xinyi Zhou, Xianzu Wang, Xinxia Shan, Y.~K. Li, Y.~Q. Wang, Y.~X. Wei, Yang Zhang, Yanhong Xu, Yao Li, Yao Zhao, Yaofeng Sun, Yaohui Wang, Yi~Yu, Yichao Zhang, Yifan Shi, Yiliang Xiong, Ying He, Yishi Piao, Yisong Wang, Yixuan Tan, Yiyang Ma, Yiyuan Liu, Yongqiang Guo, Yuan Ou, Yuduan Wang, Yue Gong, Yuheng Zou, Yujia He, Yunfan Xiong, Yuxiang Luo, Yuxiang You, Yuxuan Liu, Yuyang Zhou, Y.~X. Zhu,
  Yanhong Xu, Yanping Huang, Yaohui Li, Yi~Zheng, Yuchen Zhu, Yunxian Ma, Ying Tang, Yukun Zha, Yuting Yan, Z.~Z. Ren, Zehui Ren, Zhangli Sha, Zhe Fu, Zhean Xu, Zhenda Xie, Zhengyan Zhang, Zhewen Hao, Zhicheng Ma, Zhigang Yan, Zhiyu Wu, Zihui Gu, Zijia Zhu, Zijun Liu, Zilin Li, Ziwei Xie, Ziyang Song, Zizheng Pan, Zhen Huang, Zhipeng Xu, Zhongyu Zhang, and Zhen Zhang.
\newblock Deepseek-r1: Incentivizing reasoning capability in llms via reinforcement learning, 2025.
\newblock \url{https://arxiv.org/abs/2501.12948}.

\bibitem[Fathi et~al.(2025)Fathi, Scholak, and No{\"e}l]{fathi2025unifying}
Nima Fathi, Torsten Scholak, and Pierre-Andr{\'e} No{\"e}l.
\newblock Unifying autoregressive and diffusion-based sequence generation.
\newblock \emph{arXiv preprint arXiv:2504.06416}, 2025.

\bibitem[Feng et~al.(2025)Feng, Geng, Guan, Wu, Wang, and He]{feng2025theoretical}
Guhao Feng, Yihan Geng, Jian Guan, Wei Wu, Liwei Wang, and Di~He.
\newblock Theoretical benefit and limitation of diffusion language model.
\newblock \emph{arXiv preprint arXiv:2502.09622}, 2025.

\bibitem[Geiping et~al.(2025)Geiping, McLeish, Jain, Kirchenbauer, Singh, Bartoldson, Kailkhura, Bhatele, and Goldstein]{geiping2025scaling}
Jonas Geiping, Sean McLeish, Neel Jain, John Kirchenbauer, Siddharth Singh, Brian~R Bartoldson, Bhavya Kailkhura, Abhinav Bhatele, and Tom Goldstein.
\newblock Scaling up test-time compute with latent reasoning: A recurrent depth approach.
\newblock \emph{arXiv preprint arXiv:2502.05171}, 2025.

\bibitem[Google(2024)]{Google:GeminiThinkingMode2024}
Google.
\newblock Gemini 2.0 flash thinking mode (gemini-2.0-flash-thinking-exp-1219).
\newblock \url{https://cloud.google.com/vertex-ai/generative-ai/docs/thinking-mode}, 2024.
\newblock Accessed: 2025-04-22.

\bibitem[Google(2025{\natexlab{a}})]{Google:Gemini2.5Pro2025}
Google.
\newblock Gemini 2.5 (gemini-2.5-pro-preview).
\newblock \url{https://blog.google/technology/google-deepmind/gemini-model-thinking-updates-march-2025/}, 2025{\natexlab{a}}.
\newblock Accessed: 2025-04-22.

\bibitem[Google(2025{\natexlab{b}})]{google-gemini-2.5}
Google.
\newblock Gemini 2.5: Our most intelligent ai model, March 2025{\natexlab{b}}.
\newblock \url{https://blog.google/technology/google-deepmind/gemini-model-thinking-updates-march-2025/\#gemini-2-5-thinking}.

\bibitem[Guo et~al.(2025)Guo, Yang, Zhang, Song, Zhang, Xu, Zhu, Ma, Wang, Bi, et~al.]{guo2025deepseek}
Daya Guo, Dejian Yang, Haowei Zhang, Junxiao Song, Ruoyu Zhang, Runxin Xu, Qihao Zhu, Shirong Ma, Peiyi Wang, Xiao Bi, et~al.
\newblock Deepseek-r1: Incentivizing reasoning capability in llms via reinforcement learning.
\newblock \emph{arXiv preprint arXiv:2501.12948}, 2025.

\bibitem[Habib et~al.(2023)Habib, Fourrier, Kydlíček, Wolf, and Tunstall]{lighteval}
Nathan Habib, Clémentine Fourrier, Hynek Kydlíček, Thomas Wolf, and Lewis Tunstall.
\newblock Lighteval: A lightweight framework for llm evaluation, 2023.
\newblock \url{https://github.com/huggingface/lighteval}.

\bibitem[Hendrycks et~al.(2021)Hendrycks, Burns, Kadavath, Arora, Basart, Tang, Song, and Steinhardt]{hendrycks2021measuring}
Dan Hendrycks, Collin Burns, Saurav Kadavath, Akul Arora, Steven Basart, Eric Tang, Dawn Song, and Jacob Steinhardt.
\newblock Measuring mathematical problem solving with the math dataset.
\newblock \emph{arXiv preprint arXiv:2103.03874}, 2021.

\bibitem[Hounshell(1984)]{hounshell1984american}
David Hounshell.
\newblock \emph{From the American system to mass production, 1800-1932: The development of manufacturing technology in the United States}.
\newblock Number~4. Jhu Press, 1984.

\bibitem[Jaech et~al.(2024)Jaech, Kalai, Lerer, Richardson, El-Kishky, Low, Helyar, Madry, Beutel, Carney, et~al.]{jaech2024openai}
Aaron Jaech, Adam Kalai, Adam Lerer, Adam Richardson, Ahmed El-Kishky, Aiden Low, Alec Helyar, Aleksander Madry, Alex Beutel, Alex Carney, et~al.
\newblock Openai o1 system card.
\newblock \emph{arXiv preprint arXiv:2412.16720}, 2024.

\bibitem[Jin et~al.(2025)Jin, Cheng, Ankner, Saunshi, Elias, Yazdanbakhsh, Ragan-Kelley, Subramanian, and Carbin]{jin2025learning}
Tian Jin, Ellie~Y Cheng, Zack Ankner, Nikunj Saunshi, Blake~M Elias, Amir Yazdanbakhsh, Jonathan Ragan-Kelley, Suvinay Subramanian, and Michael Carbin.
\newblock Learning to keep a promise: Scaling language model decoding parallelism with learned asynchronous decoding.
\newblock \emph{arXiv preprint arXiv:2502.11517}, 2025.

\bibitem[Kou et~al.(2024)Kou, Hu, He, Deng, and Zhang]{kou2024cllms}
Siqi Kou, Lanxiang Hu, Zhezhi He, Zhijie Deng, and Hao Zhang.
\newblock Cllms: Consistency large language models.
\newblock In \emph{Forty-first International Conference on Machine Learning}, 2024.

\bibitem[Lou et~al.(2023)Lou, Meng, and Ermon]{lou2023discrete}
Aaron Lou, Chenlin Meng, and Stefano Ermon.
\newblock Discrete diffusion modeling by estimating the ratios of the data distribution.
\newblock \emph{arXiv preprint arXiv:2310.16834}, 2023.

\bibitem[{Mathematical Association of America}(2024)]{aime2024}
{Mathematical Association of America}.
\newblock {American Invitational Mathematics Examination 2024}, 2024.
\newblock \url{https://artofproblemsolving.com/wiki/index.php/American_Invitational_Mathematics_Examination}.
\newblock Accessed: 2025-05-14.

\bibitem[{Mathematical Association of America}(2025)]{aime2025}
{Mathematical Association of America}.
\newblock {American Invitational Mathematics Examination 2025}, 2025.
\newblock \url{https://artofproblemsolving.com/wiki/index.php/American_Invitational_Mathematics_Examination}.
\newblock Accessed: 2025-05-14.

\bibitem[McCarthy(1960)]{mccarthy1960recursive}
John McCarthy.
\newblock Recursive functions of symbolic expressions and their computation by machine, part i.
\newblock \emph{Communications of the ACM}, 3\penalty0 (4):\penalty0 184--195, 1960.

\bibitem[Muennighoff et~al.(2025)Muennighoff, Yang, Shi, Li, Fei-Fei, Hajishirzi, Zettlemoyer, Liang, Cand{\`e}s, and Hashimoto]{muennighoff2025s1}
Niklas Muennighoff, Zitong Yang, Weijia Shi, Xiang~Lisa Li, Li~Fei-Fei, Hannaneh Hajishirzi, Luke Zettlemoyer, Percy Liang, Emmanuel Cand{\`e}s, and Tatsunori Hashimoto.
\newblock s1: Simple test-time scaling.
\newblock \emph{arXiv preprint arXiv:2501.19393}, 2025.

\bibitem[Netting(1993)]{netting1993smallholders}
Robert Netting.
\newblock Smallholders, householders.
\newblock \emph{The ENVIRONMENT in anthropology: A reader in ecology, culture, and sustainable living}, 10:\penalty0 14, 1993.

\bibitem[Nie et~al.(2025)Nie, Zhu, You, Zhang, Ou, Hu, Zhou, Lin, Wen, and Li]{nie2025large}
Shen Nie, Fengqi Zhu, Zebin You, Xiaolu Zhang, Jingyang Ou, Jun Hu, Jun Zhou, Yankai Lin, Ji-Rong Wen, and Chongxuan Li.
\newblock Large language diffusion models.
\newblock \emph{arXiv preprint arXiv:2502.09992}, 2025.

\bibitem[OpenAI(2025)]{openai-o3}
OpenAI.
\newblock Introducing openai o3 and o4-mini, April 2025.
\newblock \url{https://openai.com/index/introducing-o3-and-o4-mini/}.

\bibitem[OpenAI et~al.(2024)OpenAI, :, Jaech, Kalai, Lerer, Richardson, El-Kishky, Low, Helyar, Madry, Beutel, Carney, Iftimie, Karpenko, Passos, Neitz, Prokofiev, Wei, Tam, Bennett, Kumar, Saraiva, Vallone, Duberstein, Kondrich, Mishchenko, Applebaum, Jiang, Nair, Zoph, Ghorbani, Rossen, Sokolowsky, Barak, McGrew, Minaiev, Hao, Baker, Houghton, McKinzie, Eastman, Lugaresi, Bassin, Hudson, Li, de~Bourcy, Voss, Shen, Zhang, Koch, Orsinger, Hesse, Fischer, Chan, Roberts, Kappler, Levy, Selsam, Dohan, Farhi, Mely, Robinson, Tsipras, Li, Oprica, Freeman, Zhang, Wong, Proehl, Cheung, Mitchell, Wallace, Ritter, Mays, Wang, Such, Raso, Leoni, Tsimpourlas, Song, von Lohmann, Sulit, Salmon, Parascandolo, Chabot, Zhao, Brockman, Leclerc, Salman, Bao, Sheng, Andrin, Bagherinezhad, Ren, Lightman, Chung, Kivlichan, O'Connell, Osband, Gilaberte, Akkaya, Kostrikov, Sutskever, Kofman, Pachocki, Lennon, Wei, Harb, Twore, Feng, Yu, Weng, Tang, Yu, Candela, Palermo, Parish, Heidecke, Hallman, Rizzo, Gordon, Uesato, Ward,
  Huizinga, Wang, Chen, Xiao, Singhal, Nguyen, Cobbe, Shi, Wood, Rimbach, Gu-Lemberg, Liu, Lu, Stone, Yu, Ahmad, Yang, Liu, Maksin, Ho, Fedus, Weng, Li, McCallum, Held, Kuhn, Kondraciuk, Kaiser, Metz, Boyd, Trebacz, Joglekar, Chen, Tintor, Meyer, Jones, Kaufer, Schwarzer, Shah, Yatbaz, Guan, Xu, Yan, Glaese, Chen, Lampe, Malek, Wang, Fradin, McClay, Pavlov, Wang, Wang, Murati, Bavarian, Rohaninejad, McAleese, Chowdhury, Chowdhury, Ryder, Tezak, Brown, Nachum, Boiko, Murk, Watkins, Chao, Ashbourne, Izmailov, Zhokhov, Dias, Arora, Lin, Lopes, Gaon, Miyara, Leike, Hwang, Garg, Brown, James, Shu, Cheu, Greene, Jain, Altman, Toizer, Toyer, Miserendino, Agarwal, Hernandez, Baker, McKinney, Yan, Zhao, Hu, Santurkar, Chaudhuri, Zhang, Fu, Papay, Lin, Balaji, Sanjeev, Sidor, Broda, Clark, Wang, Gordon, Sanders, Patwardhan, Sottiaux, Degry, Dimson, Zheng, Garipov, Stasi, Bansal, Creech, Peterson, Eloundou, Qi, Kosaraju, Monaco, Pong, Fomenko, Zheng, Zhou, McCabe, Zaremba, Dubois, Lu, Chen, Cha, Bai, He, Zhang, Wang,
  Shao, and Li]{openai2024openaio1card}
OpenAI, :, Aaron Jaech, Adam Kalai, Adam Lerer, Adam Richardson, Ahmed El-Kishky, Aiden Low, Alec Helyar, Aleksander Madry, Alex Beutel, Alex Carney, Alex Iftimie, Alex Karpenko, Alex~Tachard Passos, Alexander Neitz, Alexander Prokofiev, Alexander Wei, Allison Tam, Ally Bennett, Ananya Kumar, Andre Saraiva, Andrea Vallone, Andrew Duberstein, Andrew Kondrich, Andrey Mishchenko, Andy Applebaum, Angela Jiang, Ashvin Nair, Barret Zoph, Behrooz Ghorbani, Ben Rossen, Benjamin Sokolowsky, Boaz Barak, Bob McGrew, Borys Minaiev, Botao Hao, Bowen Baker, Brandon Houghton, Brandon McKinzie, Brydon Eastman, Camillo Lugaresi, Cary Bassin, Cary Hudson, Chak~Ming Li, Charles de~Bourcy, Chelsea Voss, Chen Shen, Chong Zhang, Chris Koch, Chris Orsinger, Christopher Hesse, Claudia Fischer, Clive Chan, Dan Roberts, Daniel Kappler, Daniel Levy, Daniel Selsam, David Dohan, David Farhi, David Mely, David Robinson, Dimitris Tsipras, Doug Li, Dragos Oprica, Eben Freeman, Eddie Zhang, Edmund Wong, Elizabeth Proehl, Enoch Cheung, Eric
  Mitchell, Eric Wallace, Erik Ritter, Evan Mays, Fan Wang, Felipe~Petroski Such, Filippo Raso, Florencia Leoni, Foivos Tsimpourlas, Francis Song, Fred von Lohmann, Freddie Sulit, Geoff Salmon, Giambattista Parascandolo, Gildas Chabot, Grace Zhao, Greg Brockman, Guillaume Leclerc, Hadi Salman, Haiming Bao, Hao Sheng, Hart Andrin, Hessam Bagherinezhad, Hongyu Ren, Hunter Lightman, Hyung~Won Chung, Ian Kivlichan, Ian O'Connell, Ian Osband, Ignasi~Clavera Gilaberte, Ilge Akkaya, Ilya Kostrikov, Ilya Sutskever, Irina Kofman, Jakub Pachocki, James Lennon, Jason Wei, Jean Harb, Jerry Twore, Jiacheng Feng, Jiahui Yu, Jiayi Weng, Jie Tang, Jieqi Yu, Joaquin~Quiñonero Candela, Joe Palermo, Joel Parish, Johannes Heidecke, John Hallman, John Rizzo, Jonathan Gordon, Jonathan Uesato, Jonathan Ward, Joost Huizinga, Julie Wang, Kai Chen, Kai Xiao, Karan Singhal, Karina Nguyen, Karl Cobbe, Katy Shi, Kayla Wood, Kendra Rimbach, Keren Gu-Lemberg, Kevin Liu, Kevin Lu, Kevin Stone, Kevin Yu, Lama Ahmad, Lauren Yang, Leo Liu,
  Leon Maksin, Leyton Ho, Liam Fedus, Lilian Weng, Linden Li, Lindsay McCallum, Lindsey Held, Lorenz Kuhn, Lukas Kondraciuk, Lukasz Kaiser, Luke Metz, Madelaine Boyd, Maja Trebacz, Manas Joglekar, Mark Chen, Marko Tintor, Mason Meyer, Matt Jones, Matt Kaufer, Max Schwarzer, Meghan Shah, Mehmet Yatbaz, Melody~Y. Guan, Mengyuan Xu, Mengyuan Yan, Mia Glaese, Mianna Chen, Michael Lampe, Michael Malek, Michele Wang, Michelle Fradin, Mike McClay, Mikhail Pavlov, Miles Wang, Mingxuan Wang, Mira Murati, Mo~Bavarian, Mostafa Rohaninejad, Nat McAleese, Neil Chowdhury, Neil Chowdhury, Nick Ryder, Nikolas Tezak, Noam Brown, Ofir Nachum, Oleg Boiko, Oleg Murk, Olivia Watkins, Patrick Chao, Paul Ashbourne, Pavel Izmailov, Peter Zhokhov, Rachel Dias, Rahul Arora, Randall Lin, Rapha~Gontijo Lopes, Raz Gaon, Reah Miyara, Reimar Leike, Renny Hwang, Rhythm Garg, Robin Brown, Roshan James, Rui Shu, Ryan Cheu, Ryan Greene, Saachi Jain, Sam Altman, Sam Toizer, Sam Toyer, Samuel Miserendino, Sandhini Agarwal, Santiago Hernandez,
  Sasha Baker, Scott McKinney, Scottie Yan, Shengjia Zhao, Shengli Hu, Shibani Santurkar, Shraman~Ray Chaudhuri, Shuyuan Zhang, Siyuan Fu, Spencer Papay, Steph Lin, Suchir Balaji, Suvansh Sanjeev, Szymon Sidor, Tal Broda, Aidan Clark, Tao Wang, Taylor Gordon, Ted Sanders, Tejal Patwardhan, Thibault Sottiaux, Thomas Degry, Thomas Dimson, Tianhao Zheng, Timur Garipov, Tom Stasi, Trapit Bansal, Trevor Creech, Troy Peterson, Tyna Eloundou, Valerie Qi, Vineet Kosaraju, Vinnie Monaco, Vitchyr Pong, Vlad Fomenko, Weiyi Zheng, Wenda Zhou, Wes McCabe, Wojciech Zaremba, Yann Dubois, Yinghai Lu, Yining Chen, Young Cha, Yu~Bai, Yuchen He, Yuchen Zhang, Yunyun Wang, Zheng Shao, and Zhuohan Li.
\newblock Openai o1 system card, 2024.
\newblock \url{https://arxiv.org/abs/2412.16720}.

\bibitem[Pan et~al.(2025)Pan, Li, Lian, Snell, Zhou, Yala, Darrell, Keutzer, and Suhr]{pan2025learning}
Jiayi Pan, Xiuyu Li, Long Lian, Charlie Snell, Yifei Zhou, Adam Yala, Trevor Darrell, Kurt Keutzer, and Alane Suhr.
\newblock Learning adaptive parallel reasoning with language models.
\newblock \emph{arXiv preprint arXiv:2504.15466}, 2025.

\bibitem[Qwen(2024)]{qwen2.5}
Qwen.
\newblock Qwen2: A family of open-source language models by alibaba cloud, 2024.
\newblock \url{https://github.com/QwenLM/Qwen}.

\bibitem[Qwen(2025)]{qwq32b}
Qwen.
\newblock Qwq-32b: Embracing the power of reinforcement learning, March 2025.
\newblock \url{https://qwenlm.github.io/blog/qwq-32b/}.

\bibitem[Rein et~al.(2024)Rein, Hou, Stickland, Petty, Pang, Dirani, Michael, and Bowman]{rein2024gpqa}
David Rein, Betty~Li Hou, Asa~Cooper Stickland, Jackson Petty, Richard~Yuanzhe Pang, Julien Dirani, Julian Michael, and Samuel~R Bowman.
\newblock Gpqa: A graduate-level google-proof q\&a benchmark.
\newblock In \emph{First Conference on Language Modeling}, 2024.

\bibitem[Rodionov et~al.(2025)Rodionov, Garipov, Shutova, Yakushev, Egiazarian, Sinitsin, Kuznedelev, and Alistarh]{rodionov2025hogwild}
Gleb Rodionov, Roman Garipov, Alina Shutova, George Yakushev, Vage Egiazarian, Anton Sinitsin, Denis Kuznedelev, and Dan Alistarh.
\newblock Hogwild! inference: Parallel llm generation via concurrent attention.
\newblock \emph{arXiv preprint arXiv:2504.06261}, 2025.

\bibitem[Sahoo et~al.(2024)Sahoo, Arriola, Schiff, Gokaslan, Marroquin, Chiu, Rush, and Kuleshov]{sahoo2024simple}
Subham Sahoo, Marianne Arriola, Yair Schiff, Aaron Gokaslan, Edgar Marroquin, Justin Chiu, Alexander Rush, and Volodymyr Kuleshov.
\newblock Simple and effective masked diffusion language models.
\newblock \emph{Advances in Neural Information Processing Systems}, 37:\penalty0 130136--130184, 2024.

\bibitem[Shi et~al.(2024)Shi, Han, Wang, Doucet, and Titsias]{shi2024simplified}
Jiaxin Shi, Kehang Han, Zhe Wang, Arnaud Doucet, and Michalis Titsias.
\newblock Simplified and generalized masked diffusion for discrete data.
\newblock \emph{Advances in neural information processing systems}, 37:\penalty0 103131--103167, 2024.

\bibitem[Vaswani et~al.(2017)Vaswani, Shazeer, Parmar, Uszkoreit, Jones, Gomez, Kaiser, and Polosukhin]{vaswani2017attention}
Ashish Vaswani, Noam Shazeer, Niki Parmar, Jakob Uszkoreit, Llion Jones, Aidan~N Gomez, {\L}ukasz Kaiser, and Illia Polosukhin.
\newblock Attention is all you need.
\newblock \emph{Advances in neural information processing systems}, 30, 2017.

\bibitem[Wang et~al.(2025)Wang, Schiff, Sahoo, and Kuleshov]{wang2025remasking}
Guanghan Wang, Yair Schiff, Subham~Sekhar Sahoo, and Volodymyr Kuleshov.
\newblock Remasking discrete diffusion models with inference-time scaling.
\newblock \emph{arXiv preprint arXiv:2503.00307}, 2025.

\bibitem[Wang et~al.(2022)Wang, Wei, Schuurmans, Le, Chi, Narang, Chowdhery, and Zhou]{wang2022self}
Xuezhi Wang, Jason Wei, Dale Schuurmans, Quoc Le, Ed~Chi, Sharan Narang, Aakanksha Chowdhery, and Denny Zhou.
\newblock Self-consistency improves chain of thought reasoning in language models.
\newblock \emph{arXiv preprint arXiv:2203.11171}, 2022.

\bibitem[xAI(2025)]{grok-3}
xAI.
\newblock Grok 3 beta — the age of reasoning agents, February 2025.
\newblock \url{https://x.ai/news/grok-3}.

\bibitem[Yang et~al.(2025)Yang, Chen, and Chen]{yang2025ape}
Xinyu Yang, Tianqi Chen, and Beidi Chen.
\newblock Ape: Faster and longer context-augmented generation via adaptive parallel encoding.
\newblock \emph{arXiv preprint arXiv:2502.05431}, 2025.

\bibitem[Yao et~al.(2023)Yao, Yu, Zhao, Shafran, Griffiths, Cao, and Narasimhan]{yao2023tree}
Shunyu Yao, Dian Yu, Jeffrey Zhao, Izhak Shafran, Tom Griffiths, Yuan Cao, and Karthik Narasimhan.
\newblock Tree of thoughts: Deliberate problem solving with large language models.
\newblock \emph{Advances in neural information processing systems}, 36:\penalty0 11809--11822, 2023.

\bibitem[Ye et~al.(2025{\natexlab{a}})Ye, Xie, Zheng, Gao, Wu, Jiang, Li, and Kong]{dream2025}
Jiacheng Ye, Zhihui Xie, Lin Zheng, Jiahui Gao, Zirui Wu, Xin Jiang, Zhenguo Li, and Lingpeng Kong.
\newblock Dream 7b, 2025{\natexlab{a}}.
\newblock \url{https://hkunlp.github.io/blog/2025/dream}.

\bibitem[Ye et~al.(2025{\natexlab{b}})Ye, Huang, Xiao, Chern, Xia, and Liu]{ye2025limo}
Yixin Ye, Zhen Huang, Yang Xiao, Ethan Chern, Shijie Xia, and Pengfei Liu.
\newblock Limo: Less is more for reasoning.
\newblock \emph{arXiv preprint arXiv:2502.03387}, 2025{\natexlab{b}}.

\bibitem[Zhang et~al.(2024)Zhang, Huang, Zhou, Li, and Ouyang]{zhang2024accessing}
Di~Zhang, Xiaoshui Huang, Dongzhan Zhou, Yuqiang Li, and Wanli Ouyang.
\newblock Accessing gpt-4 level mathematical olympiad solutions via monte carlo tree self-refine with llama-3 8b.
\newblock \emph{arXiv preprint arXiv:2406.07394}, 2024.

\bibitem[Zhao et~al.(2025)Zhao, Gupta, Zheng, and Grover]{zhao2025d1}
Siyan Zhao, Devaansh Gupta, Qinqing Zheng, and Aditya Grover.
\newblock d1: Scaling reasoning in diffusion large language models via reinforcement learning.
\newblock \emph{arXiv preprint arXiv:2504.12216}, 2025.

\bibitem[Zheng et~al.(2023)Zheng, Yin, Xie, Huang, Sun, Yu, Cao, Kozyrakis, Stoica, Gonzalez, et~al.]{zheng2023efficiently}
Lianmin Zheng, Liangsheng Yin, Zhiqiang Xie, Jeff Huang, Chuyue Sun, Cody\_Hao Yu, Shiyi Cao, Christos Kozyrakis, Ion Stoica, Joseph~E Gonzalez, et~al.
\newblock Efficiently programming large language models using sglang.
\newblock 2023.

\end{thebibliography}
